\documentclass[runningheads]{llncs}
\usepackage{booktabs}
\usepackage{graphicx}
\usepackage{multirow}
\usepackage{subfig}
\begin{document}
%
\title{Small, but important: Traffic light proposals for detecting small traffic lights and beyond}
\titlerunning{Traffic light proposals for detecting small traffic lights and beyond}
%
%
\author{Tom Sanitz\inst{1,2,\ast}, Christian Wilms\inst{2,\ast}, and Simone Frintrop\inst{2}}
\authorrunning{T. Sanitz et al.}
%
\institute{Ibeo Automotive Systems GmbH, Hamburg, Germany\\ \and Computer Vision Group, University of Hamburg, Hamburg, Germany
\email{\{christian.wilms,simone.frintrop\}@uni-hamburg.de}\\ \ \\
$\ast$ indicates equal contribution
}
\maketitle              
%
\begin{abstract}
Traffic light detection is a challenging problem in the context of self-driving cars and driver assistance systems. While most existing systems produce good results on large traffic lights, detecting small and tiny ones is often overlooked. A key problem here is the inherent downsampling in CNNs, leading to low-resolution features for detection. To mitigate this problem, we propose a new traffic light detection system, comprising a novel traffic light proposal generator that utilizes findings from general object proposal generation, fine-grained multi-scale features, and attention for efficient processing. Moreover, we design a new detection head for classifying and refining our proposals. We evaluate our system on three challenging, publicly available datasets and compare it against six methods. The results show substantial improvements of at least $12.6\%$ on small and tiny traffic lights, as well as strong results across all sizes of traffic lights.
\end{abstract}
\section{Introduction}

Traffic light detection, which involves locating traffic lights and classifying their state, is an essential task for self-driving cars and driver assistance systems. Due to the complex nature of urban environments with several intersections, heavy traffic, and distracting objects, detecting traffic lights is challenging~\cite{kim2018deep}. Moreover, safe driving in such environments is challenging to humans as well, leading to stress, oversights, and potentially fatal accidents~\cite{jensen2016vision}. Hence, support by driver assistance systems is of great importance in such environments.

\begin{figure}[tb]
\centering
\begin{tabular}{ccc}
\subfloat{\includegraphics[width = .31\linewidth]{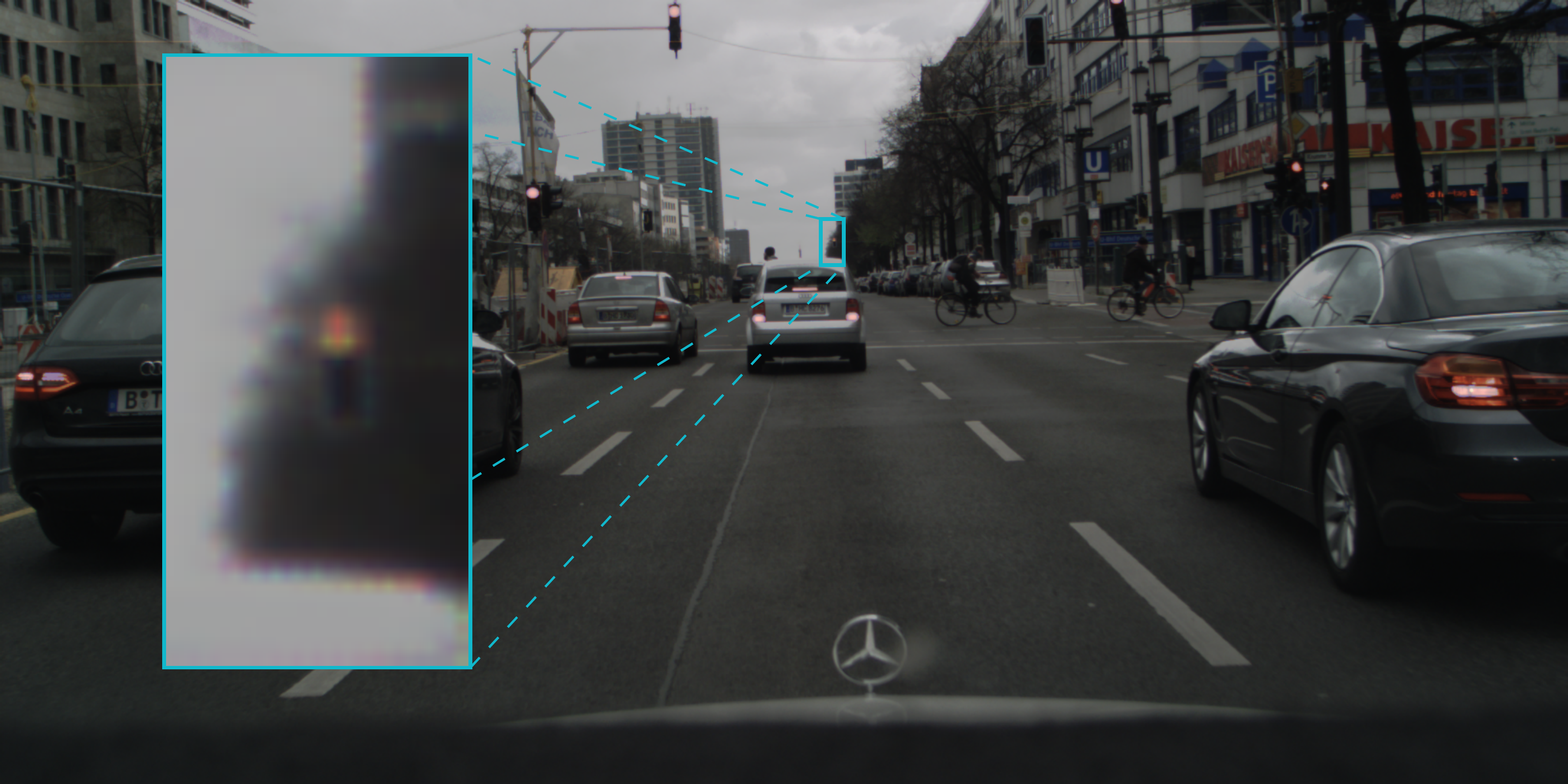}} &
\subfloat{\includegraphics[width = .31\linewidth]{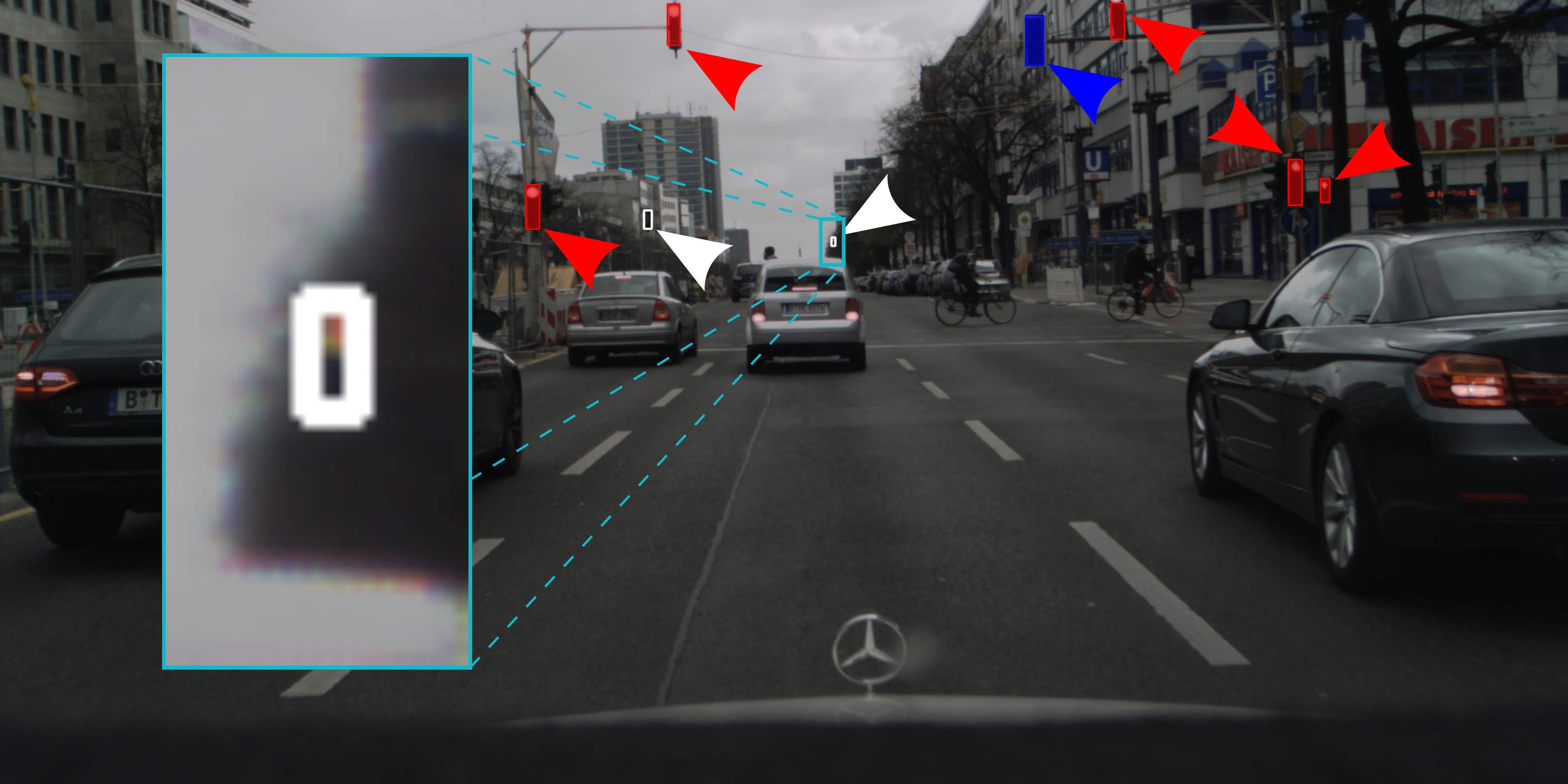}} &
\subfloat{\includegraphics[width = .31\linewidth]{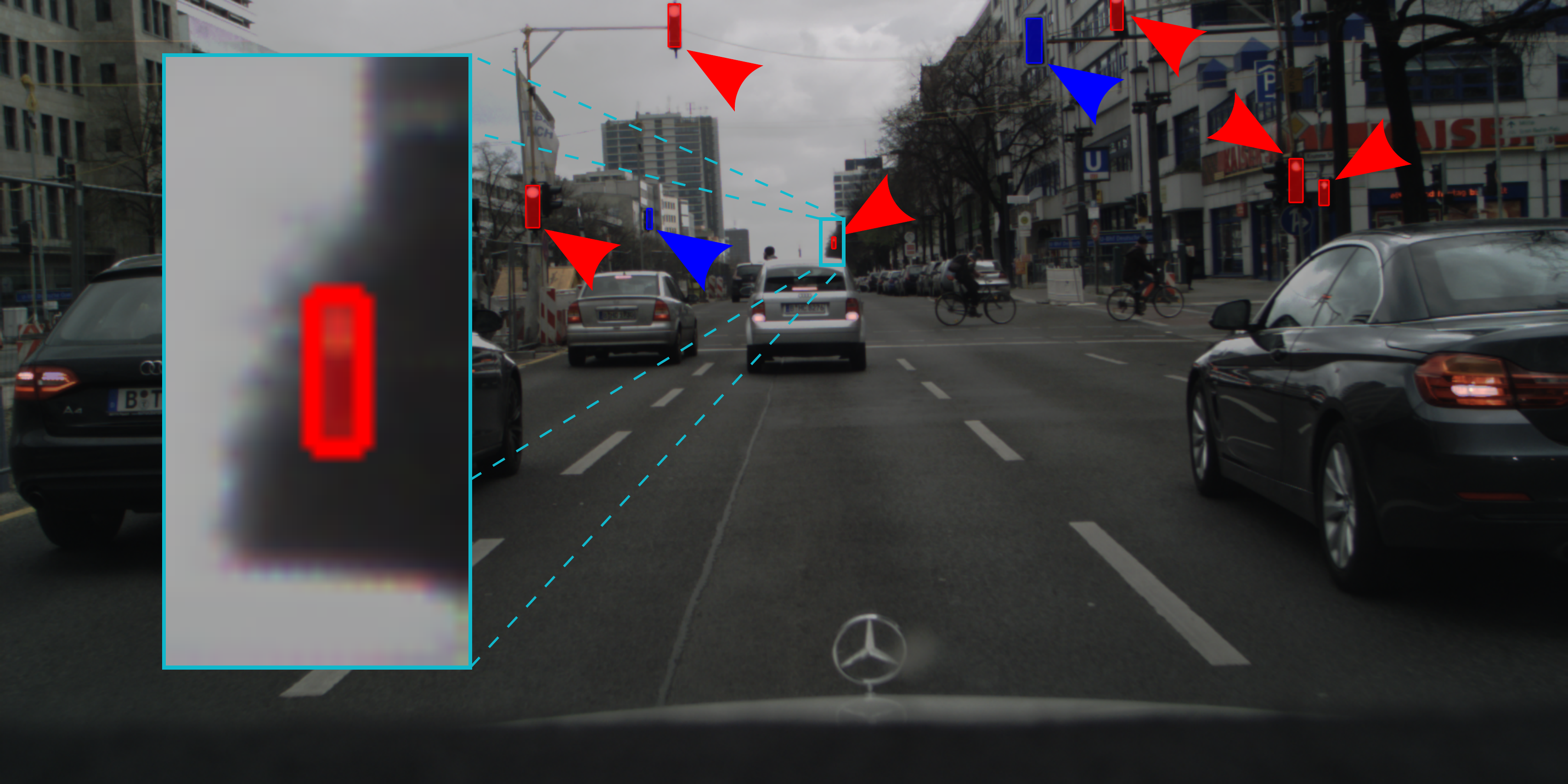}}\\
Image & TL-SSD~\cite{muller2018detecting} & Ours
\end{tabular}
\caption{Qualitative results of TL-SSD~\cite{muller2018detecting} and our proposed traffic light detection system, which highlight the weak performance of current systems on small and tiny traffic lights. The color indicates the assigned traffic light state. Dark blue denotes the state \textit{off}, while white denotes missed or misclassified traffic lights. Arrows highlight traffic lights for better visibility.}
\label{fig:qualResIntro}
\end{figure}

Several approaches for traffic light detection were proposed, mostly based on standard object detectors~\cite{behrendt2017deep,bach2018deep,pon2018hierarchical,muller2018detecting}.  However, the detection of traffic lights appearing small or tiny on the image plane remains a problem.  A major reason is the inherent subsampling in CNNs to extract semantically rich features. Moreover, small and tiny traffic lights are much smaller than objects typically annotated in object detection datasets like COCO~\cite{lin2014microsoft}. Hence, standard object detectors might not be well-suited for the problem of detecting such traffic lights, as visible in the results in Fig.~\ref{fig:qualResIntro}. Despite the difficulties, small and tiny traffic lights are essential for safe, efficient, and eco-friendly driving. For instance, detecting traffic lights early, when they are still far away and  small or tiny on the image plane, allows a car to slowly approach a traffic light showing a stop signal. This reduces noise emission, fuel consumption, and carbon emission~\cite{wu2011fuel,rittger2015driving}.

Several directions were proposed to improve the detection of small and tiny objects in computer vision literature~\cite{chen2020survey}. Prominent directions include tiling approaches~\cite{bargoti2017deep,ozge2019power,wilms2021airline,wilms2022localizing}, which process sub-images to increase the objects' relative size. However, this leads to substantially increased runtimes~\cite{wilms2022localizing}. Another direction is a multi-scale feature representation, combing semantically rich, low-resolution features with high-resolution features lacking high-level semantics~\cite{gong2021effective,lin2017feature}. A third direction are object proposal generators to discover small objects~\cite{hu2017finding,WilmsFrintropACCV2018,lu2018toward}. Since the latter two directions do not suffer from substantially increased runtime, we follow them to improve the detection of small and tiny traffic lights.

In this paper, we propose a new traffic light detection system, focusing on small and tiny traffic lights, while also improving the detection across all sizes of traffic lights. First, to better localize small and tiny traffic lights, we propose a new traffic light proposal generator. It utilizes the one-shot paradigm from object proposal generation~\cite{Hu2017-fastmask}, introduces an improved  multi-scale feature representation to create semantically rich features at high resolution, and employs attention for increased efficiency. To classify and refine the proposals, we adapt the Faster R-CNN framework~\cite{ren2015faster} utilizing our high-quality proposals and a new traffic light detection head. Our extensive evaluation on three challenging, publicly available datasets demonstrates the effectiveness of our approach for detecting small and tiny traffic lights, as well as traffic lights of all sizes. Across all three datasets, we outperform all other tested methods and strong baselines.

Overall, our contributions are threefold:
\begin{itemize}
    \item A novel traffic light detection system focusing on small and tiny traffic lights.
    \item A new traffic light proposal generator using a multi-scale feature representation and attention.
    \item An extensive evaluation on three challenging datasets, outperforming all other tested methods and strong baselines across all datasets.
\end{itemize}


\section{Related Work}
Traffic light detection has attracted considerable attention over the last decade. While early approaches used traditional techniques including color thresholding or extracting shape information, see~\cite{jensen2016vision} for a survey, CNNs have been utilized recently. Subsequently, we review the most important and relevant approaches utilizing RGB data only.

Most recent CNN-based traffic light detection approaches extend standard object detectors like SSD~\cite{liu2016ssd}, YOLO~\cite{redmon2016you}, or Faster R-CNN~\cite{ren2015faster}. For instance, \cite{muller2018detecting}~and \cite{kim2018deep} improve SSD's detection quality for small traffic lights by proposing modified anchors or an additional classifier. 
Various versions of YOLO are also adapted for traffic light detection~\cite{behrendt2017deep,lee2019accurate,wang2022traffic}. To address the challenge of tiny traffic lights, \cite{behrendt2017deep}~use a tiling approach, while \cite{wang2022traffic}~combine features from various layers of the network for detection. Finally, Faster R-CNN is commonly utilized for traffic light detection with adjustments to the anchor boxes~\cite{bach2018deep}, backbones~\cite{bach2018deep,kim2018efficient}, color spaces~\cite{kim2018efficient}, and the classification head~\cite{bach2018deep,pon2018hierarchical,gupta2019framework}.

Apart from approaches based on standard object detectors, few notable approaches exist. \cite{ouyang2019deep}~use a fully convolutional approach with a hierarchical per-pixel classifier. \cite{weber2018hdtlr}~propose a heuristics-based proposal generator to improve the detection of small traffic lights. 
Recently, \cite{yang2022scrdet++}~introduced a detection method for small objects that disentangles the features of different classes and instances, which is also applied to traffic light detection.

Overall, for the challenging detection of small and tiny traffic lights, most approaches use variations of standard object detectors or utilize heuristics-based approaches. In contrast, we propose a new traffic light detection system with a dedicated traffic light proposal generator that introduces an improved  multi-scale feature representation to create semantically rich features at high resolution as well as attention for increased efficiency.

\section{Method}
\begin{figure}[tb]
\centering
\includegraphics[width = \linewidth]{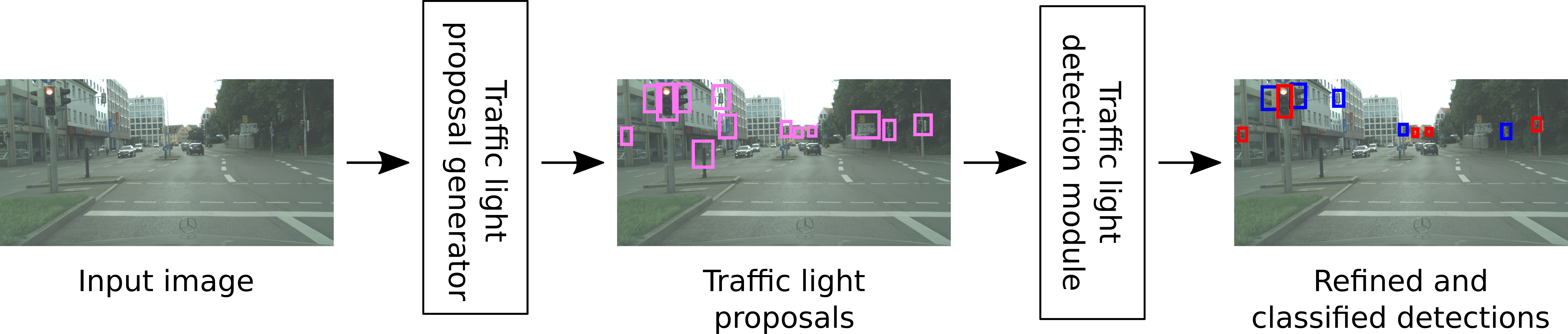}
\caption{Overview of our traffic light detection system. First, our novel traffic light proposal generator creates traffic light proposals (pink boxes) covering possible traffic light locations. Subsequently, our detection module refines and classifies each proposal (red/blue boxes).}
\label{fig:absSysFig}
\end{figure}

This section introduces our new traffic light detection system, visualized in Fig.~\ref{fig:absSysFig}. Given an input image, our novel traffic light proposal generator, described in Sec.~\ref{sec:method_TLPG}, extracts possible traffic light locations. The proposal generator explicitly addresses the challenging localization of small traffic lights by improving the resolution of the multi-scale feature representation and utilizes attention for efficient processing. Subsequently, our traffic light detection module refines the proposals and assigns them to a traffic light state based on a new traffic light detection head~(see Sec.~\ref{sec:method_TLDM}). Finally, we discuss the training of our system in Sec~\ref{sec:method_train}.

\subsection{Traffic Light Proposal Generator}
\label{sec:method_TLPG}

\begin{figure}[tb]
\centering
\includegraphics[width = \linewidth]{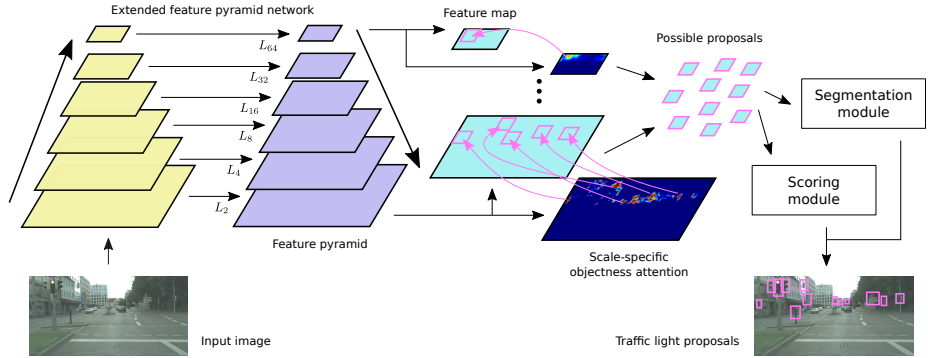}
\caption{Detailed view of our traffic light proposal generator. Our extended feature pyramid network (yellow and violet pyramids) extracts semantically rich features across all levels/resolutions ($L_2$ to $L_{64}$), yielding a multi-scale feature pyramid~(violet pyramid). For each level, we generate a feature map~(large cyan boxes) and a scale-specific objectness attention map highlighting possible traffic light locations. Subsequently, we extract windows~(possible proposals) at high-attention locations across the feature pyramid. Finally, a segmentation and a score are generated per window to create a ranked list of traffic light proposals.}
\label{fig:sysFig}
\end{figure}

The first stage of our traffic light detection system consists of our novel traffic light proposal generator, visualized in Fig.~\ref{fig:sysFig}. It is designed to locate all traffic lights and reduce the search space for the subsequent traffic light detection module. Similar to recent object proposal generators~\cite{Hu2017-fastmask,WilmsFrintropACCV2018,wilms2021superpixel}, we follow the efficient one-shot approach~\cite{Hu2017-fastmask}. Hence, the input image is processed only once by the backbone network, yielding a feature representation that is further subsampled to create a multi-scale feature pyramid. By extracting fixed-size windows from the feature pyramid, proposals for objects of different sizes are generated.

However, simply following this approach is insufficient for traffic light detection with many tiny objects. The additional subsampling to create a feature pyramid on top of the backbone leads to low-resolution feature maps and missing tiny objects. To circumvent this problem, we use the backbone itself as a multi-scale feature pyramid, introducing less subsampling. Yet, two issues arise.

First, the features from early layers of backbones like ResNets~\cite{he2016deep} are not semantically rich. To generate a feature pyramid that is semantically rich across all levels, we employ and extend the Feature Pyramid Network~(FPN)~\cite{lin2017feature} as backbone~(yellow and violet pyramids in Fig.~\ref{fig:sysFig}). The FPN combines a typical CNN-backbone with top-down and lateral connections to create a semantically rich representation across all levels (resolutions) of the feature pyramid~(violet pyramid in Fig.~\ref{fig:sysFig}). In Fig.~\ref{fig:sysFig}, the levels are denoted by $L_n$ with the downsampling factor $n$. We further add a new level to the feature pyramid for tiny objects ($L_2$ in Fig.~\ref{fig:sysFig}). This multi-scale feature pyramid serves as the base for our proposal generator.

The second issue is the large number of possible locations for extracting windows, i.e., possible traffic light proposals, in our feature pyramid due to the high resolution. To address this issue, we utilize scale-specific objectness attention proposed by~\cite{WilmsFrintropACCV2018}. The attention maps are learned per feature pyramid level~(scale-specific) and focus the window extraction as well as further processing on the most relevant areas, omitting the background. 

Based on the attention maps, we extract all relevant, i.e., high attention, windows of size $10\times 10$ across all levels of our feature pyramid~(small cyan boxes in Fig.~\ref{fig:sysFig}). Subsequently, we use the common head structure of~\cite{Hu2017-fastmask} to score the possible proposals and generate a segmentation per proposal. The final result of the traffic light proposal generator are $n$ traffic light proposals with bounding box coordinates and a score to process only the most relevant ones.

\subsection{Traffic Light Detection Module}
\label{sec:method_TLDM}

To assign the proposals generated by our traffic light proposal generator to the traffic light states and refine their locations, we apply our traffic light detection module, which is inspired by the Faster R-CNN architecture~\cite{ren2015faster}. First, we extract a feature representation of the image utilizing an FPN backbone. Note that we can share this backbone with the proposal generator.

Given the feature representation of the image and our proposals, we apply region of interest pooling per proposal, leading to a $7 \times 7$ feature map per proposal. Subsequently, each proposal is processed by our new traffic light detection head consisting of four fully-connected layers with 2048 neurons each. Based on these features, the detection head refines the traffic light location and assigns a traffic light state or the background label to each proposal.





\subsection{Training}
\label{sec:method_train}
This section describes the training of our traffic light detection system. First, we train our traffic light proposal generator utilizing the annotated data of the respective dataset, omitting any state information. We follow~\cite{Hu2017-fastmask,WilmsFrintropACCV2018} for defining the respective loss functions of the attention modules and the head for segmenting as well as scoring the traffic light proposals. 

After training the traffic light proposal generator, we utilize it to generate positive and negative training samples for our traffic light detection module. A proposal is regarded as a positive sample if it has an Intersection over Union~(IoU) of at least 0.5 with any annotated traffic light, while proposals with an IoU below 0.3 are regarded as negative samples. The rest of the training regime is similar to~\cite{ren2015faster}. 


\section{Experiments}
\label{sec:eval}
We evaluate our approach on three challenging, publicly available datasets and compare it to six systems across the different datasets. The datasets are the Bosch Small Traffic Lights Dataset~(BSTLD)~\cite{behrendt2017deep}, 
the DriveU Traffic Light Da\-ta\-set~(DTLD)~\cite{fregin2018driveu}, and the recently published dataset Cityscapes TL++~(CS-TL)~\cite{janosovits2022cityscapes}. They consist of 1978 to 40953 images with an average of 2.4 to 6.6 annotated traffic lights per image. All images show traffic scenes captured from the perspective of the driver and are of high quality. The number of traffic light states differs between the datasets. While all datasets include the states \textit{stop}, \textit{warning}, and \textit{go}, DTLD has \textit{stop/warning} annotated and all datasets include the states \textit{off} or \textit{unknown}. Note that we adapted all systems to the respective number of traffic light states contained in a dataset.

We compare our approach to different methods per dataset due to the limited availability of results and code as well as varying dataset splits. On BSTLD, we compare to~\cite{behrendt2017deep}, \cite{pon2018hierarchical}, and the SSD and Faster R-CNN baselines\footnote{Results taken from \url{https://github.com/bosch-ros-pkg/bstld}} provided by~\cite{behrendt2017deep}. 
For DTLD, we generate the results of~\cite{muller2018detecting} with their publicly available system\footnote{\url{https://github.com/julimueller/tl\_ssd}}. Since no results  are publicly available yet on the CS-TL dataset, we compare to a strong baseline using Faster R-CNN~\cite{ren2015faster} with an FPN backbone~\cite{lin2017feature}. 

To assess the quality of traffic light detection results, we use the mean Average Precision (mAP) with an IoU of 0.5 as the main measure, following~\cite{bach2018deep,pon2018hierarchical,lee2019accurate,yang2022scrdet++}. For a more detailed analysis, we report state-specific and size-specific results. While the traffic light states are fixed by the datasets, we define four relative size ranges, based on the ratio of an annotated traffic light's area and the image area. The ranges for relative size $a$ are tiny~($a\leq0.01\%$), small~($0.01\% < a\leq0.03\%$), medium~($0.03\% < a\leq0.05\%$), and large~($0.05\% <  a$), denoted as mAP$_\mathit{T}$, mAP$_\mathit{S}$, mAP$_\mathit{M}$, and mAP$_\mathit{L}$. The ranges are determined such that across all datasets, each class comprises between $20\%$ and $30\%$ of the annotated traffic lights. Finally, we also report  mAP$_\mathit{weighted}$ on BSTLD, defined by~\cite{behrendt2017deep}, which incorporates the distribution of traffic light states in the dataset.

The subsequent sections discuss the quantitative results, selected qualitative results, and two ablation studies justifying design choices.

\subsection{Quantitative Results}
\label{sec:evalQuan}
The quantitative results on BSTLD~(see Tab.~\ref{tab:boschQuanResults}), 
DTLD~(see Tab.~\ref{tab:driveuQuanResults}), and the CS-TL dataset~(see Tab.~\ref{tab:csQuanResults}) all show similar trends.  Across all datasets, our proposed system outperforms all other methods and baselines in terms of mAP. Within the datasets, the results for the individual traffic light states vary. One reason is the imbalance of annotations in the datasets. For instance, on DTLD, the states \textit{stop} and \textit{go} amount to almost $85\%$ of the annotations in the dataset. Hence, the results for those states are substantially better than for the other states. Another reason is the intra-class diversity of the states \textit{off} and \textit{unknown}. 

On BSTLD, the results in Tab.~\ref{tab:boschQuanResults} show that our proposed traffic light detection system outperforms both SSD TLD and Faster R-CNN NAS~\cite{behrendt2017deep} based on standard object detectors by $39.5\%$ and $33.0\%$ in terms of mAP. For the state \textit{warning}, the improvement is even up to $159\%$. Similarly, in terms of mAP$_\mathit{weighted}$, our system outperforms all other methods by up to $97.2\%$. 

The results on DTLD in Tab.~\ref{tab:driveuQuanResults} comparing TL-SSD~\cite{muller2018detecting} based on a standard object detector to our proposed system show again a strong improvement across all traffic light states ($+67.8\%$). On the recently published CS-TL dataset, we compare to the strong Faster R-CNN+FPN~\cite{ren2015faster,lin2017feature} baseline~(see Tab.~\ref{tab:csQuanResults}). While across all states, i.e. overall, our traffic light detection system outperforms the baseline ($+1\%$), the per-state results favor both systems twice. 


\begin{table}[tb]
\centering
\caption{Traffic light detection results on the BSTLD test set in terms of mAP, mAP$_\mathit{weighted}$, and state-specific mAPs.}
\label{tab:boschQuanResults}
\begin{tabular}{@{}lcccccc@{}}
\toprule
System & mAP$_\mathit{weighted}$ & mAP & mAP$_\mathit{stop}$ & mAP$_\mathit{warning}$ & mAP$_\mathit{go}$ & mAP$_\mathit{off}$ \\ \midrule
YOLO TLD~\cite{behrendt2017deep}  & 0.360 & - & - & - & - & - \\ 
HDA~\cite{pon2018hierarchical} & 0.530 & - & - & - & - & - \\ 
SSD TLD~\cite{behrendt2017deep} & 0.600 & 0.410 & 0.550 & 0.410 & 0.680 & \textbf{0.000} \\
Faster R-CNN NAS~\cite{behrendt2017deep} & 0.650 & 0.430 & 0.660 & 0.330 & 0.710 & \textbf{0.000} \\
Ours & \textbf{0.710} & \textbf{0.572} & \textbf{0.678} & \textbf{0.855} & \textbf{0.753} & \textbf{0.000} \\
\bottomrule
\end{tabular}
\end{table}


\begin{table}[tb]
\centering
\caption{Traffic light detection results on the DTLD test set in terms of mAP and state-specific mAPs.}
\label{tab:driveuQuanResults}
\begin{tabular}{@{}lcccccc@{}}
\toprule
System & mAP & mAP$_\mathit{stop}$ & mAP$_\mathit{stop/warning}$& mAP$_\mathit{warning}$ & mAP$_\mathit{go}$ & mAP$_\mathit{off}$ \\ \midrule
TL-SSD~\cite{muller2018detecting} & 0.329 & 0.439 & 0.283 & 0.167 & 0.583 & 0.010 \\
Ours & \textbf{0.552} & \textbf{0.699} & \textbf{0.564} & \textbf{0.564} & \textbf{0.789} & \textbf{0.142} \\
\bottomrule
\end{tabular}
\end{table}

\begin{table}[tb]
\centering
\caption{Traffic light detection results on the CS-TL test set in terms of mAP and state-specific mAPs.}
\label{tab:csQuanResults}
\begin{tabular}{@{}lccccc@{}}
\toprule
System & mAP & mAP$_\mathit{stop}$ & mAP$_\mathit{warning}$ & mAP$_\mathit{go}$ & mAP$_\mathit{unknown}$ \\ \midrule
Faster R-CNN+FPN~\cite{ren2015faster,lin2017feature} & 0.496 & \textbf{0.645} & \textbf{0.381} & 0.626 & 0.332 \\
Ours & \textbf{0.500} & 0.638 & 0.338 & \textbf{0.650} & \textbf{0.375} \\
\bottomrule
\end{tabular}
\end{table}

Analyzing the detection performance in more detail, Tab.~\ref{tab:quanResultsSize} shows the results for different relative sizes of traffic lights. Across all systems and datasets, the results on tiny traffic lights are substantially worse compared to the three other size ranges, with a drop of up to $91\%$. Note that the traffic light in BSTLD are smaller, resulting in worse results compared to the other datasets. Due to the lack of publicly available results and the novel and custom nature of the evaluation, we can only compare to TL-SSD on DTLD and the Faster R-CNN+FPN baseline on the CS-TL dataset. On both datasets, we outperform the other methods on tiny traffic lights in terms of mAP$_\mathit{T}$ ($+262\%$ and $+12.6\%$). This confirms the strong performance of our system on such traffic lights. Across the other size ranges, we also outperform TL-SSD on DTLD by an average of $79.7\%$. On the CS-TL dataset, we outperform the strong Faster R-CNN+FPN baseline on small and medium traffic lights, with a slight drop on larger ones. 

\begin{table}[tb]
\centering
\caption{Size-specific traffic light detection results on the test splits of BSTLD, DTLD, and the CS-TL dataset. $T$, $S$, $M$, and $L$ denote traffic light size ranges tiny, small, medium, and large.}
\label{tab:quanResultsSize}
\begin{tabular}{@{}rlcccc@{}}
\toprule
Dataset & System & mAP$_\mathit{T}$ & mAP$_\mathit{S}$ & mAP$_\mathit{M}$ & mAP$_\mathit{L}$ \\ \midrule
BSTLD & Ours &  0.061 & 0.508 & 0.518 & 0.679 \\ \midrule 
\multirow{2}{*}{DTLD} & TL-SSD~\cite{muller2018detecting} & 0.079 & 0.299 & 0.331 & 0.530 \\ 
& Ours &  \textbf{0.286} & \textbf{0.625} & \textbf{0.672} & \textbf{0.674} \\  \midrule 
\multirow{2}{*}{CS-TL}  & Faster R-CNN+FPN~\cite{ren2015faster,lin2017feature} & 0.207 & 0.387 & 0.385 & \textbf{0.608} \\ 
& Ours &  \textbf{0.233} & \textbf{0.421} & \textbf{0.400} & 0.531 \\
\bottomrule
\end{tabular}
\end{table}

Overall, the quantitative results across three datasets show the strong performance of our traffic light detection system. Particularly, the size-specific results indicate the substantially improved detection of small and tiny traffic lights.

\begin{figure}[tb]
\centering
\begin{tabular}{ccc}
\subfloat{\includegraphics[width = .31\linewidth]{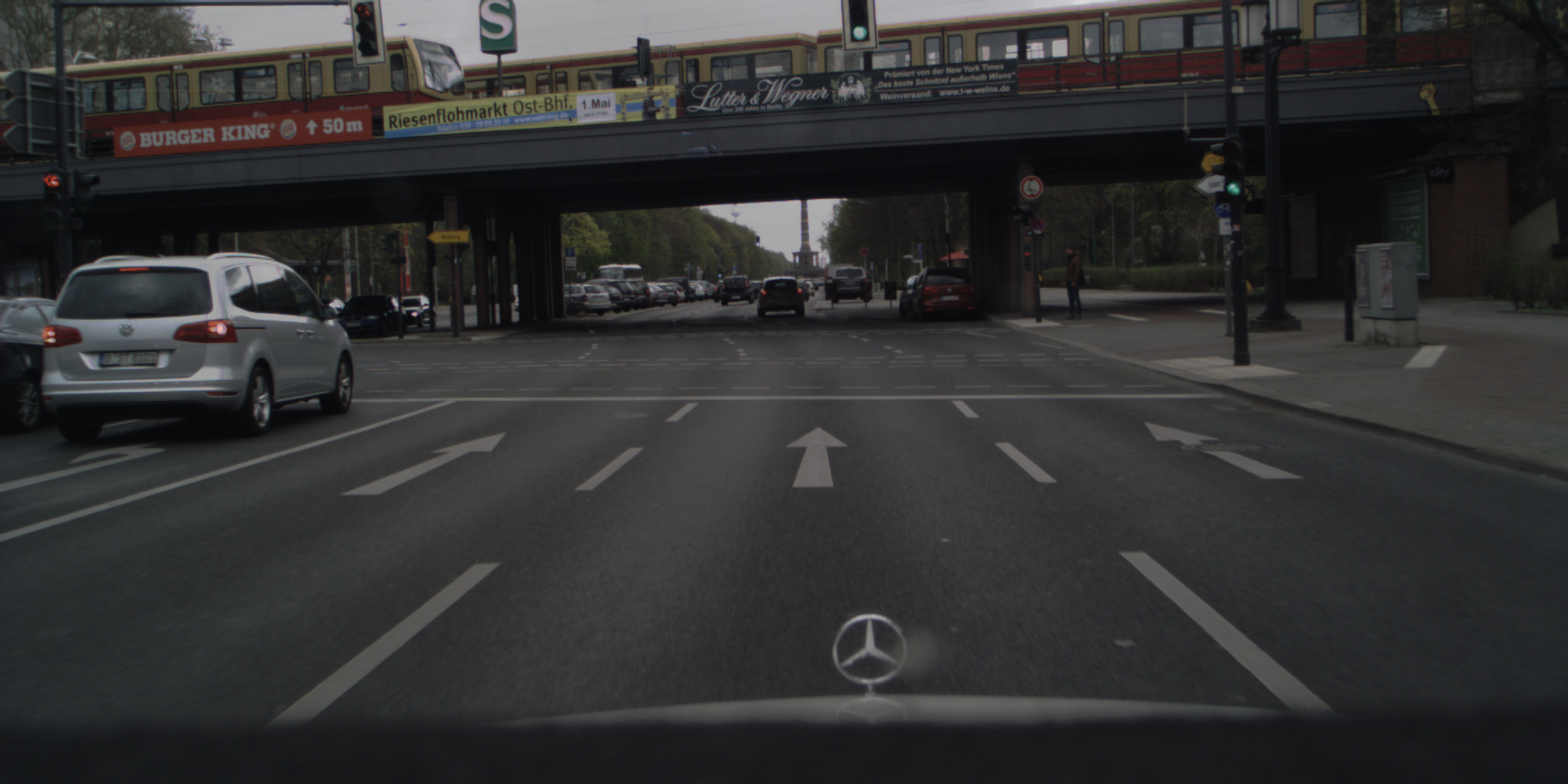}} &
\subfloat{\includegraphics[width = .31\linewidth]{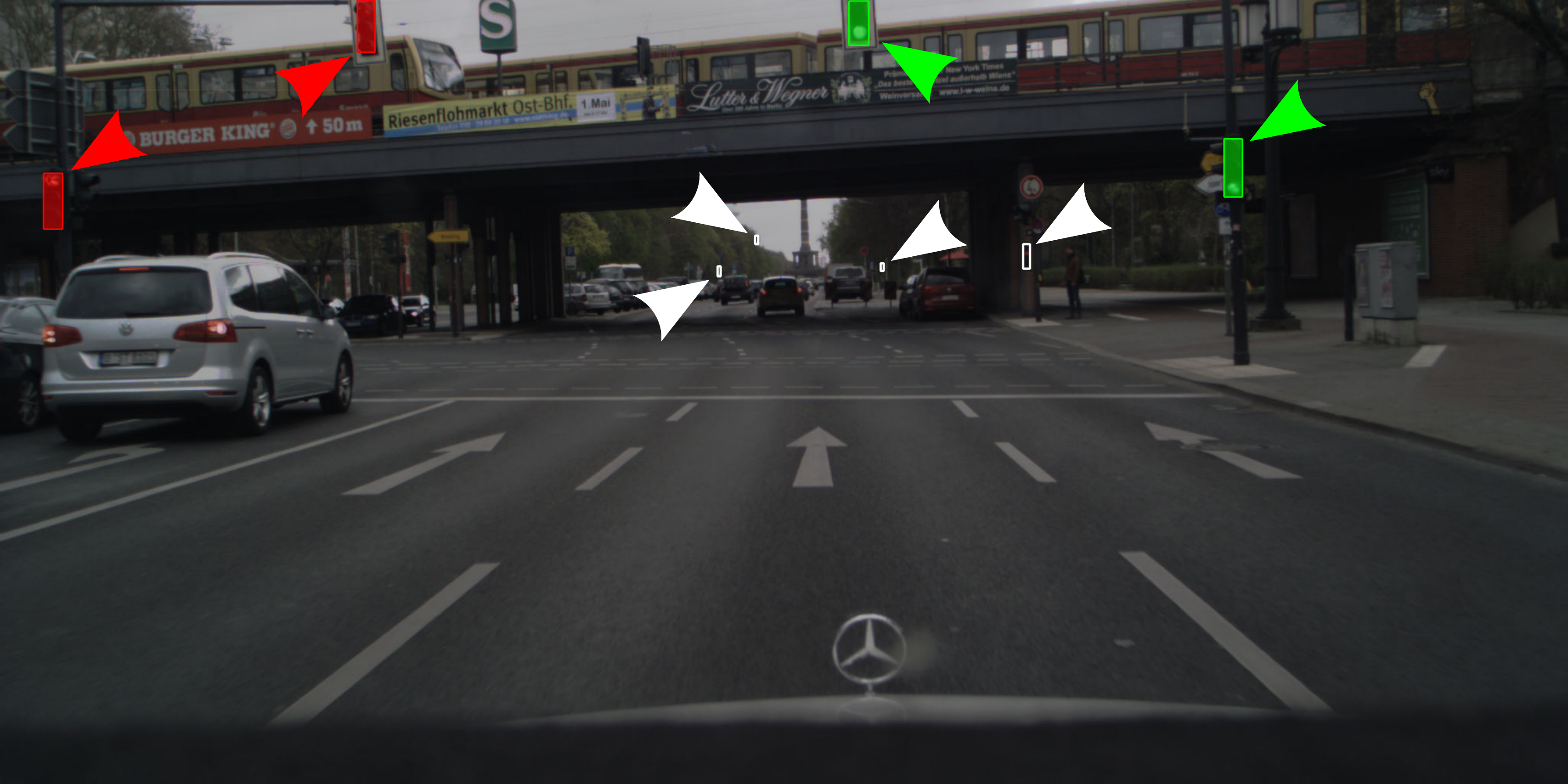}} &
\subfloat{\includegraphics[width = .31\linewidth]{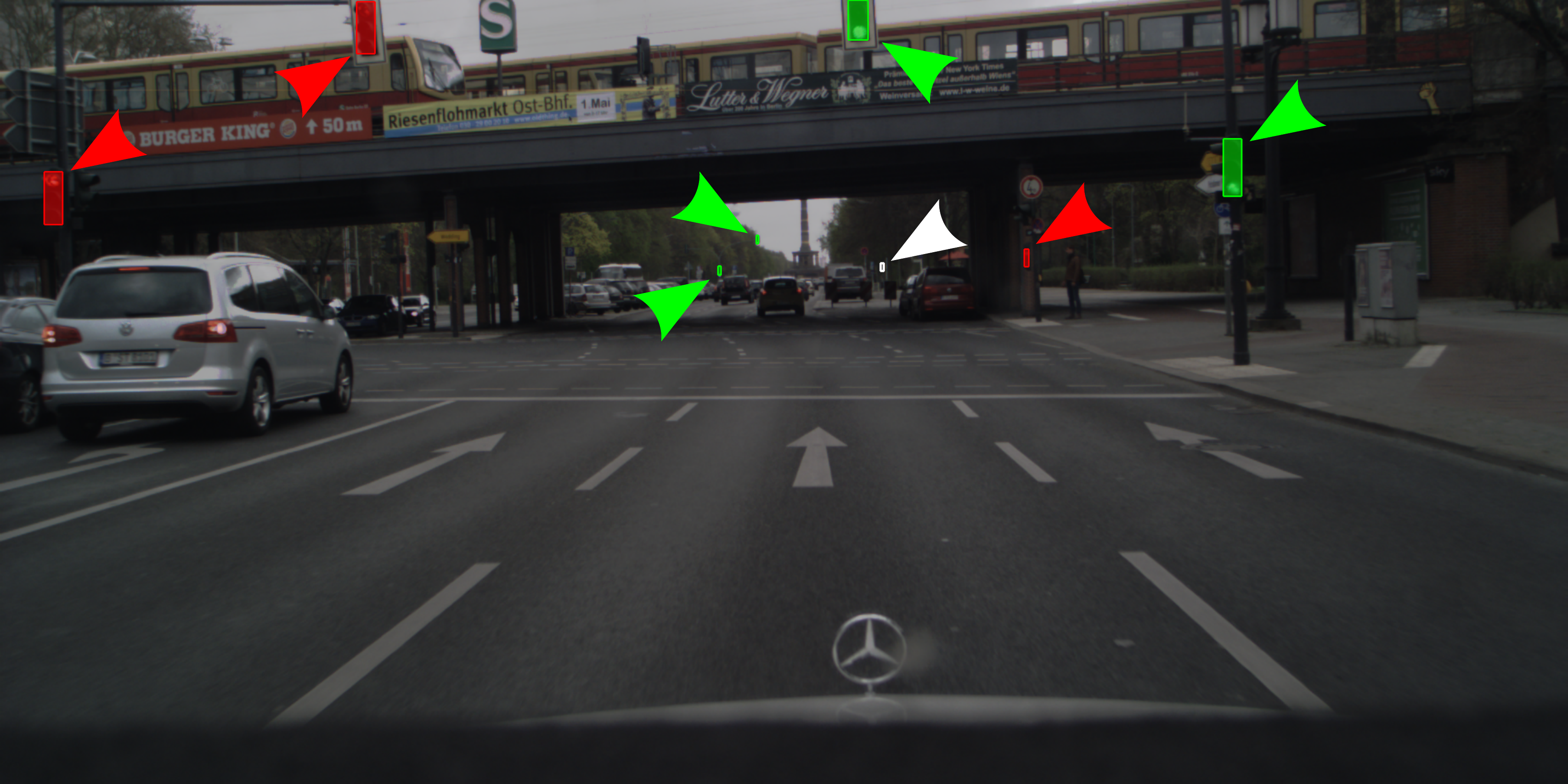}}\\
\subfloat{\includegraphics[width = .31\linewidth]{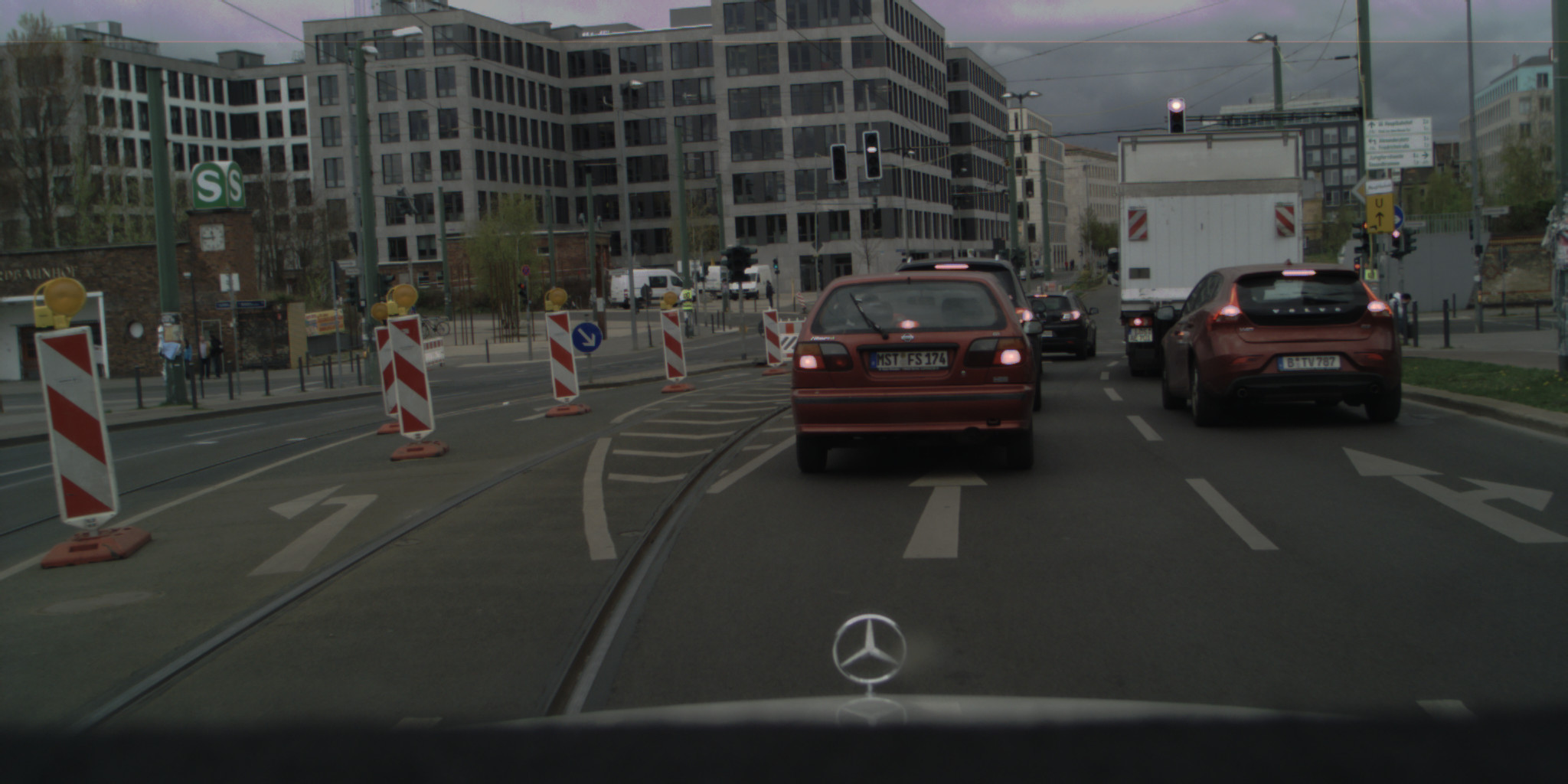}} &
\subfloat{\includegraphics[width = .31\linewidth]{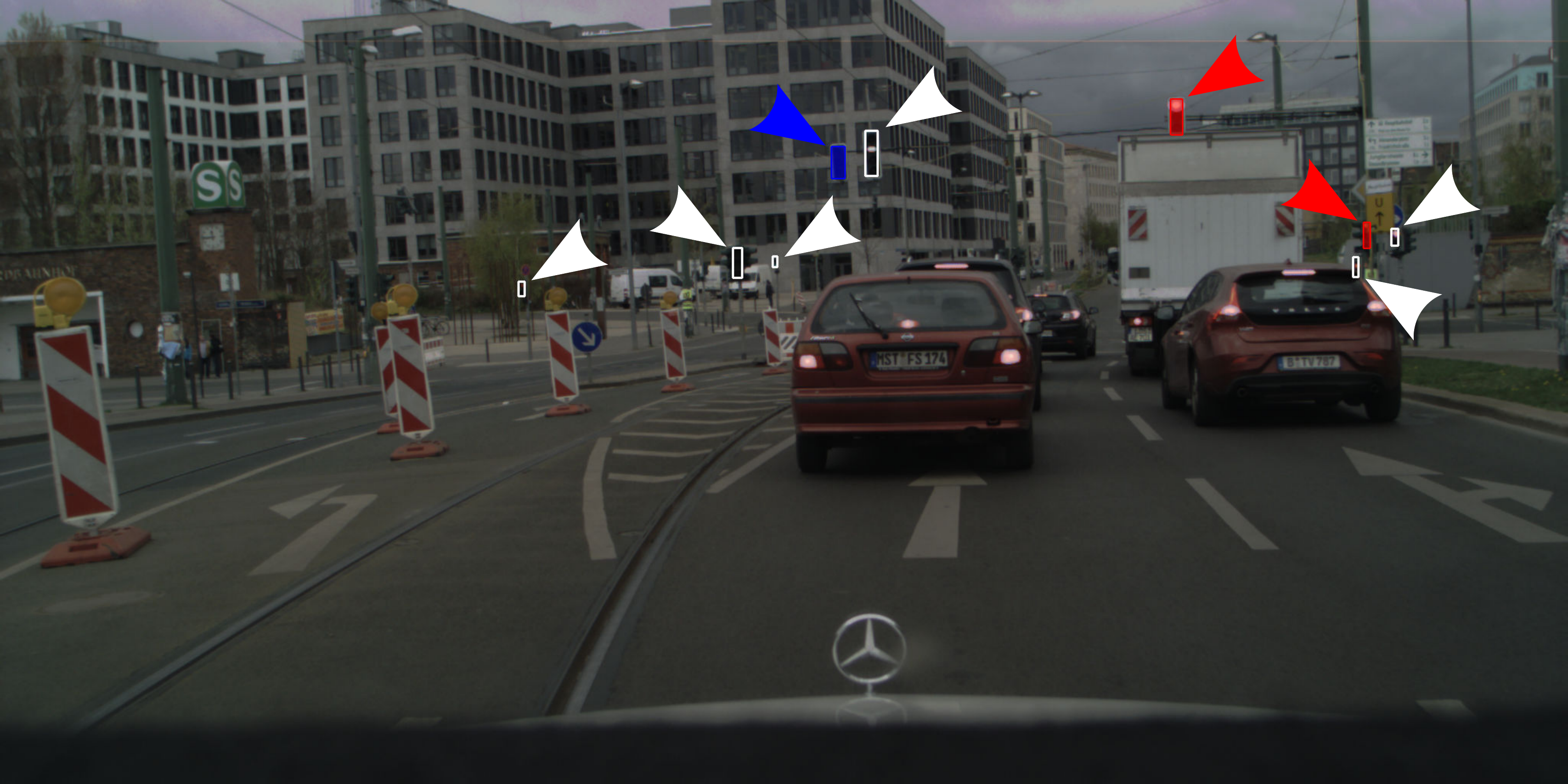}} &
\subfloat{\includegraphics[width = .31\linewidth]{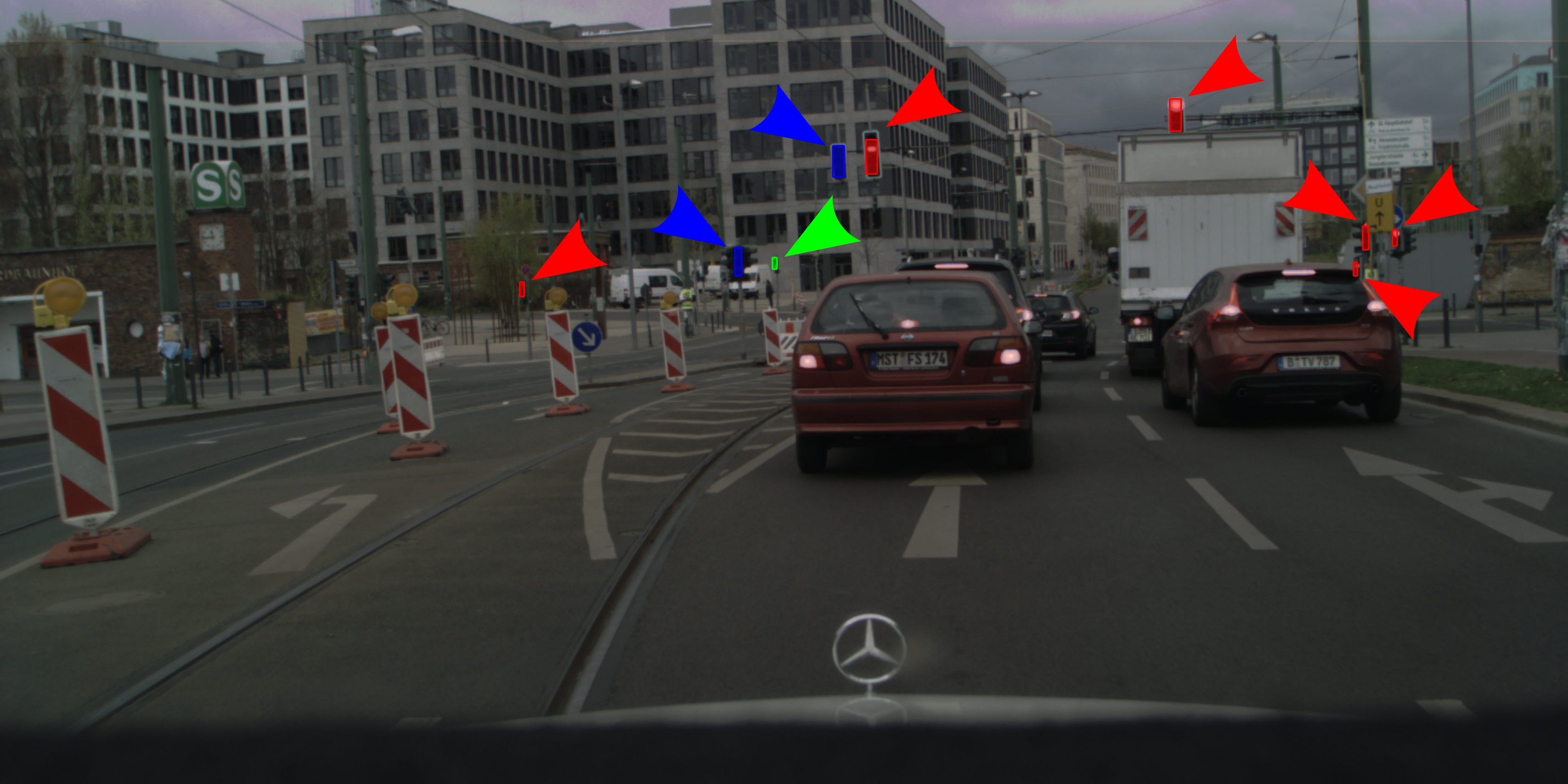}}\\

Image & TL-SSD~\cite{muller2018detecting} & Ours
\end{tabular}
\caption{Qualitative results of TL-SSD~\cite{muller2018detecting} and our proposed traffic light detection system on the DTLD test set. The color indicates the assigned traffic light state. Blue denotes the state \textit{off}, while white denotes missed or misclassified traffic lights. Arrows highlight traffic lights for better visibility.}
\label{fig:qualResDriveU}
\end{figure}

\begin{figure}[tb]
\centering
\begin{tabular}{ccc}
\subfloat{\includegraphics[width = .31\linewidth]{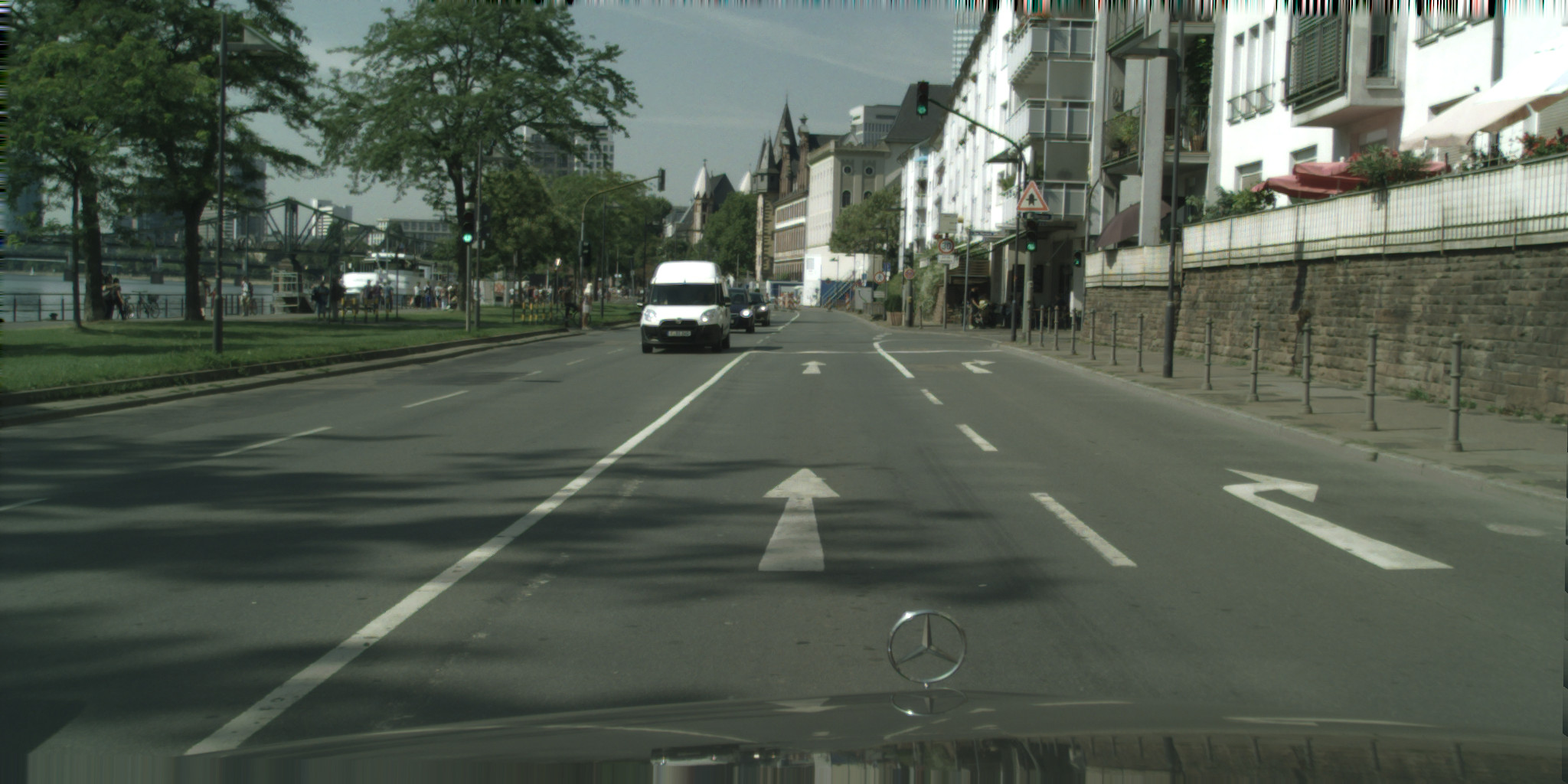}} &
\subfloat{\includegraphics[width = .31\linewidth]{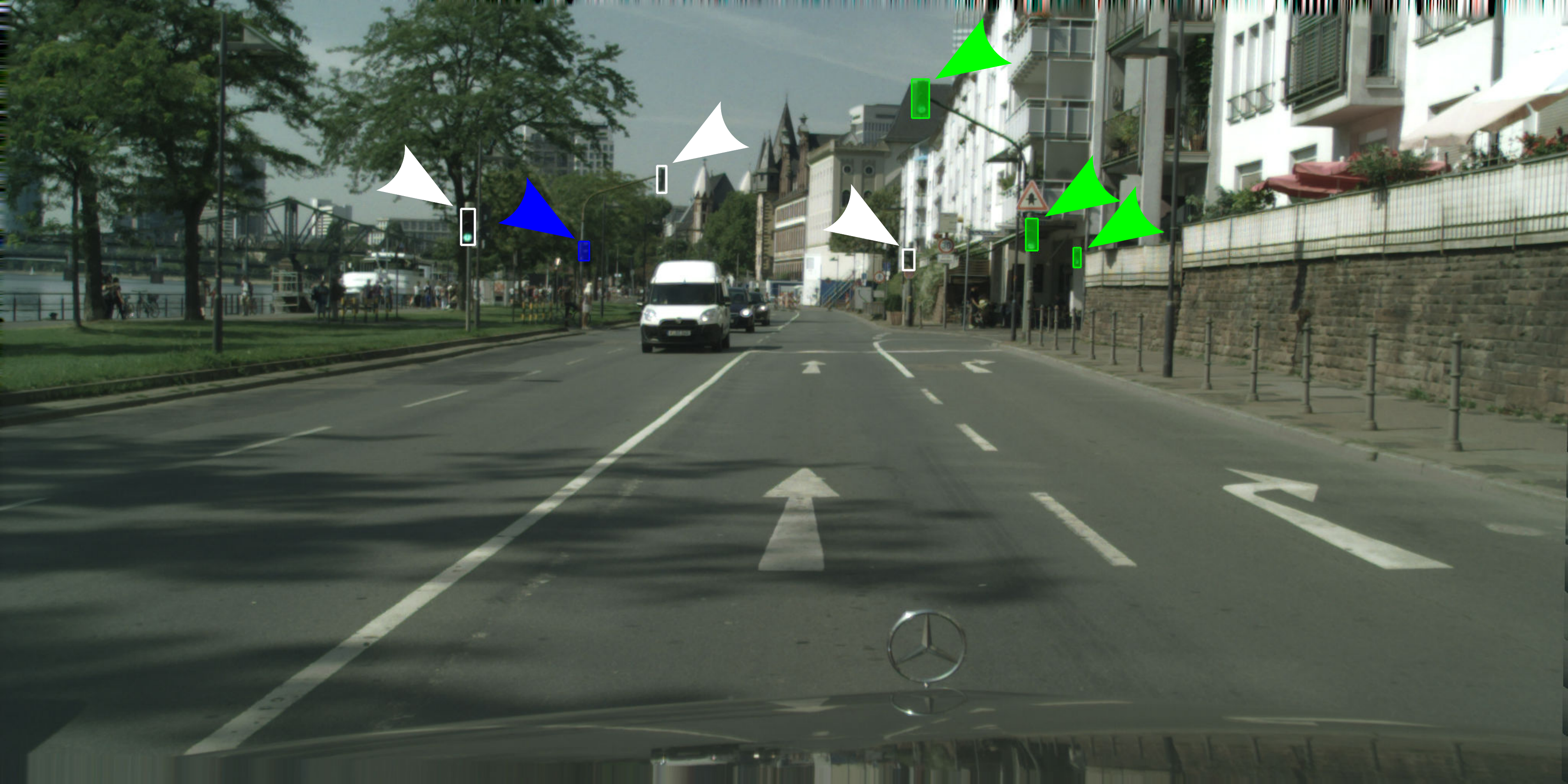}} &
\subfloat{\includegraphics[width = .31\linewidth]{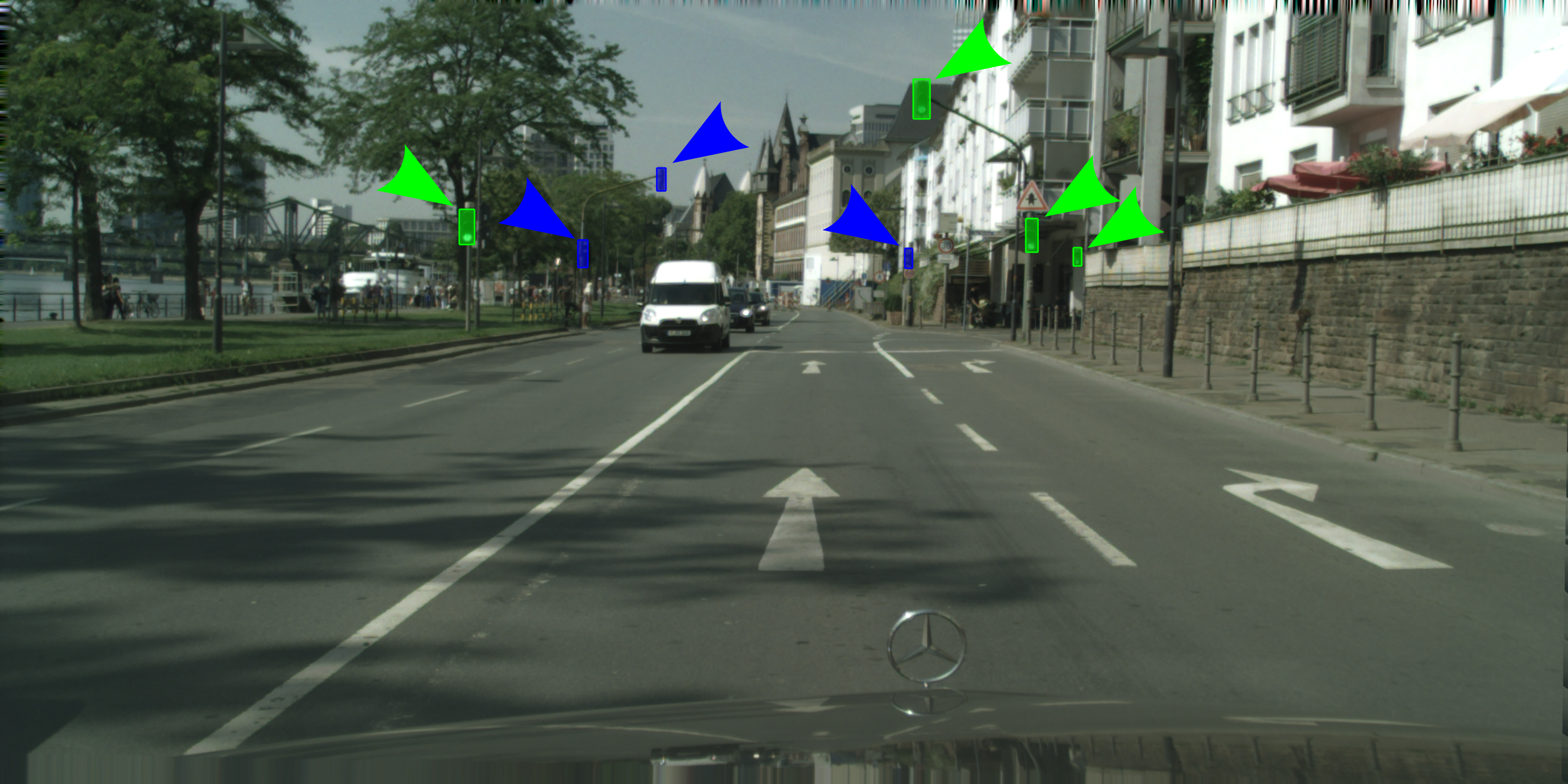}}\\
\subfloat{\includegraphics[width = .31\linewidth]{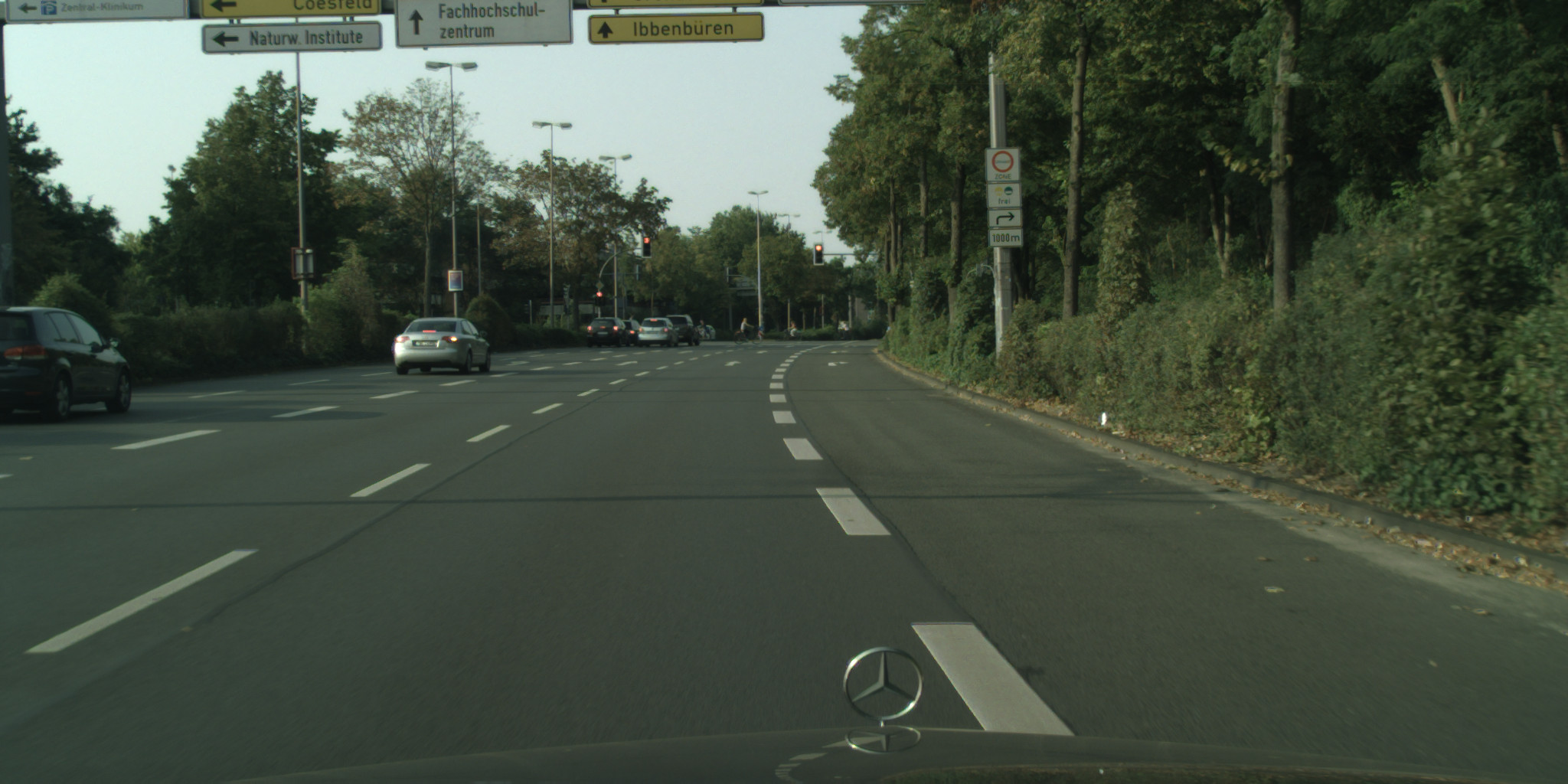}} &
\subfloat{\includegraphics[width = .31\linewidth]{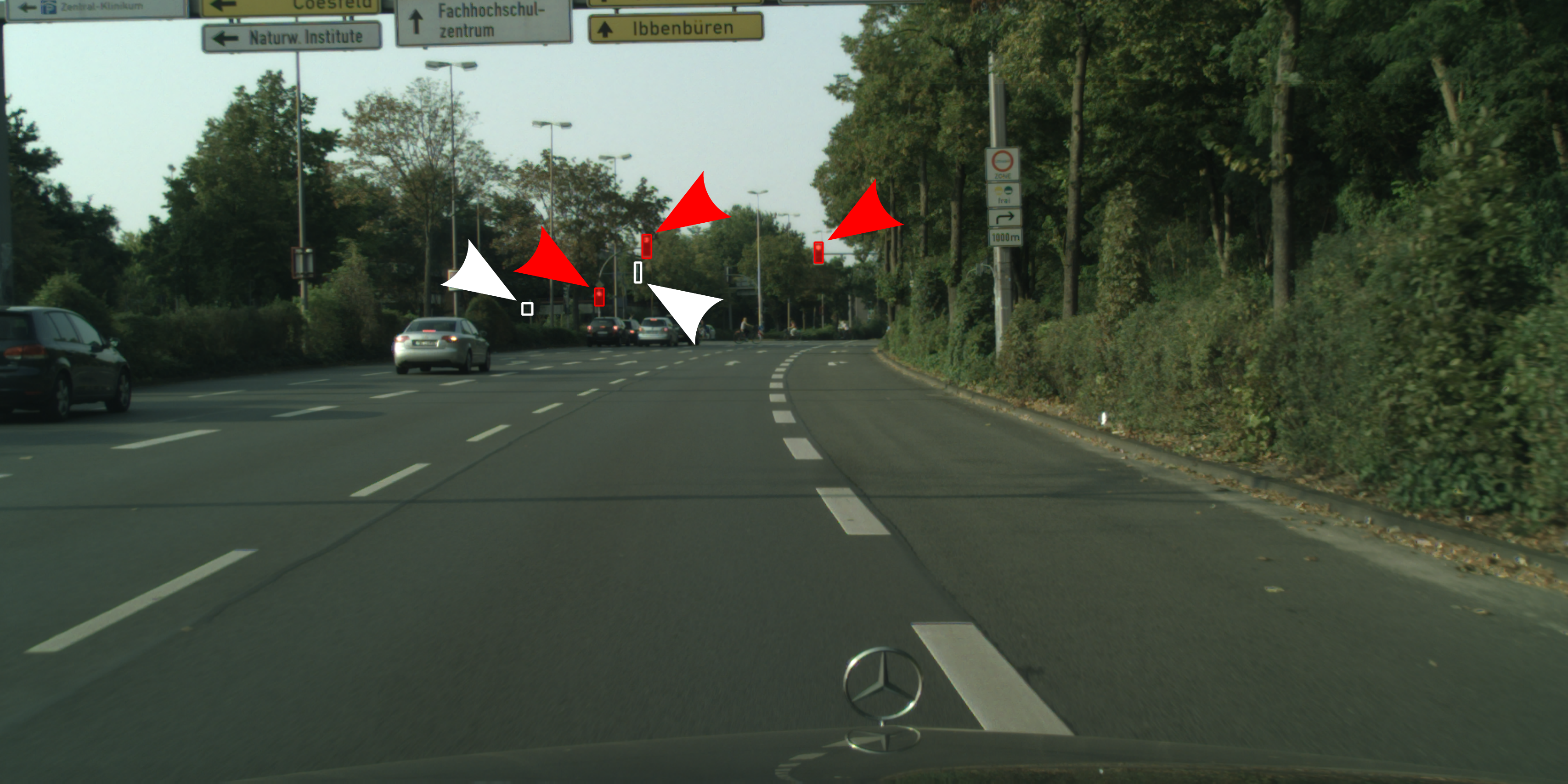}} &
\subfloat{\includegraphics[width = .31\linewidth]{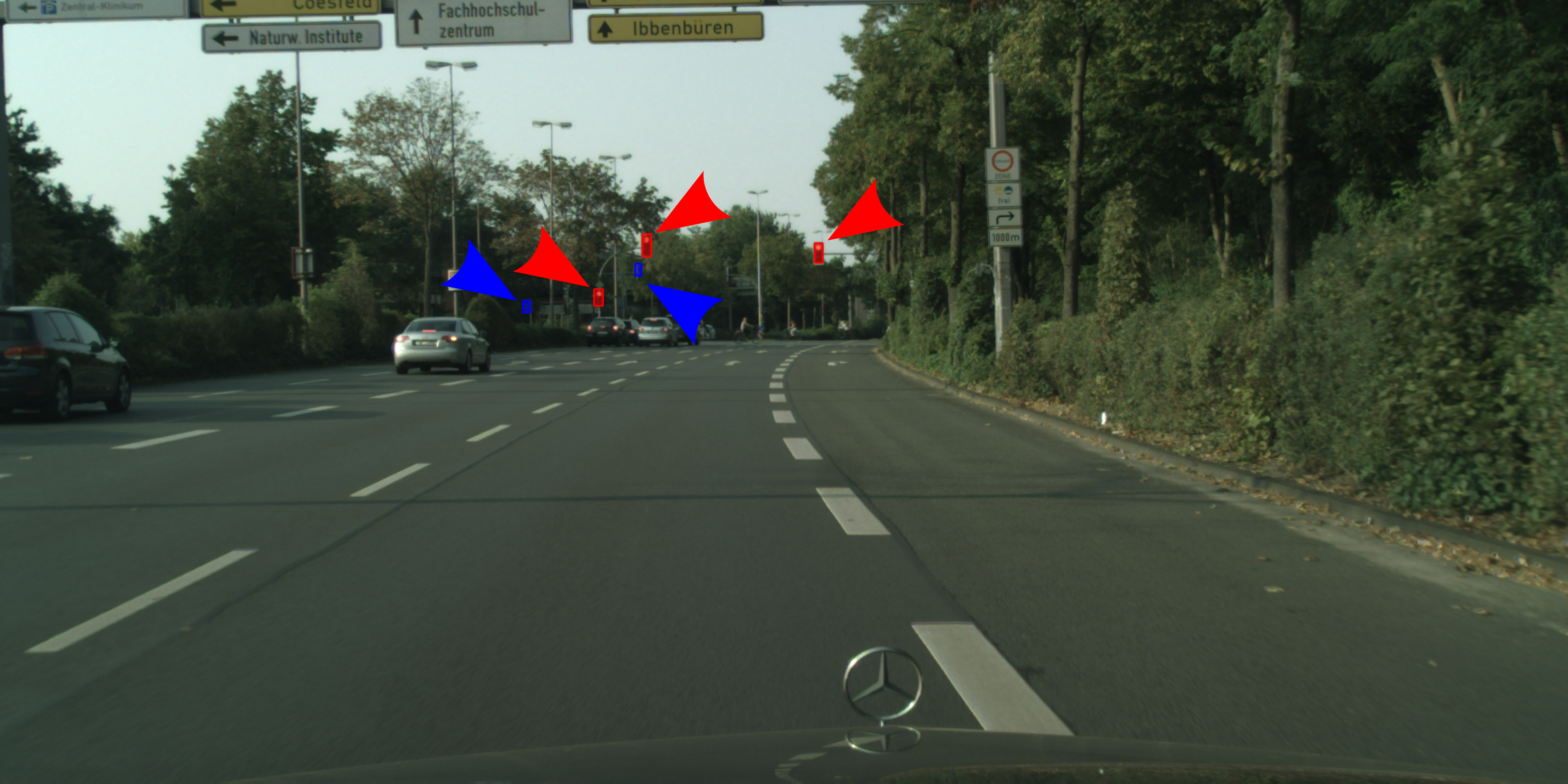}}\\
\subfloat{\includegraphics[width = .31\linewidth]{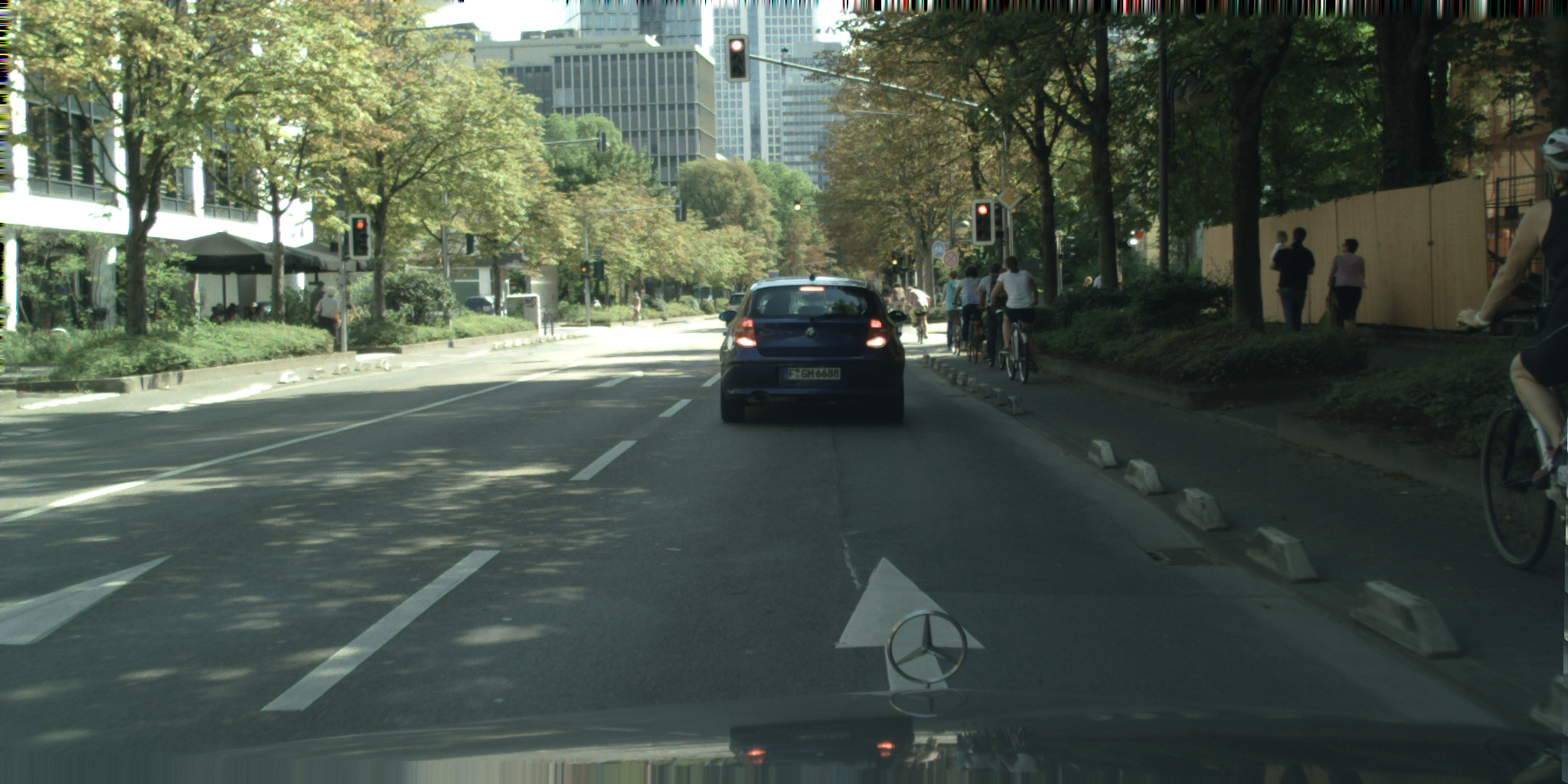}} &
\subfloat{\includegraphics[width = .31\linewidth]{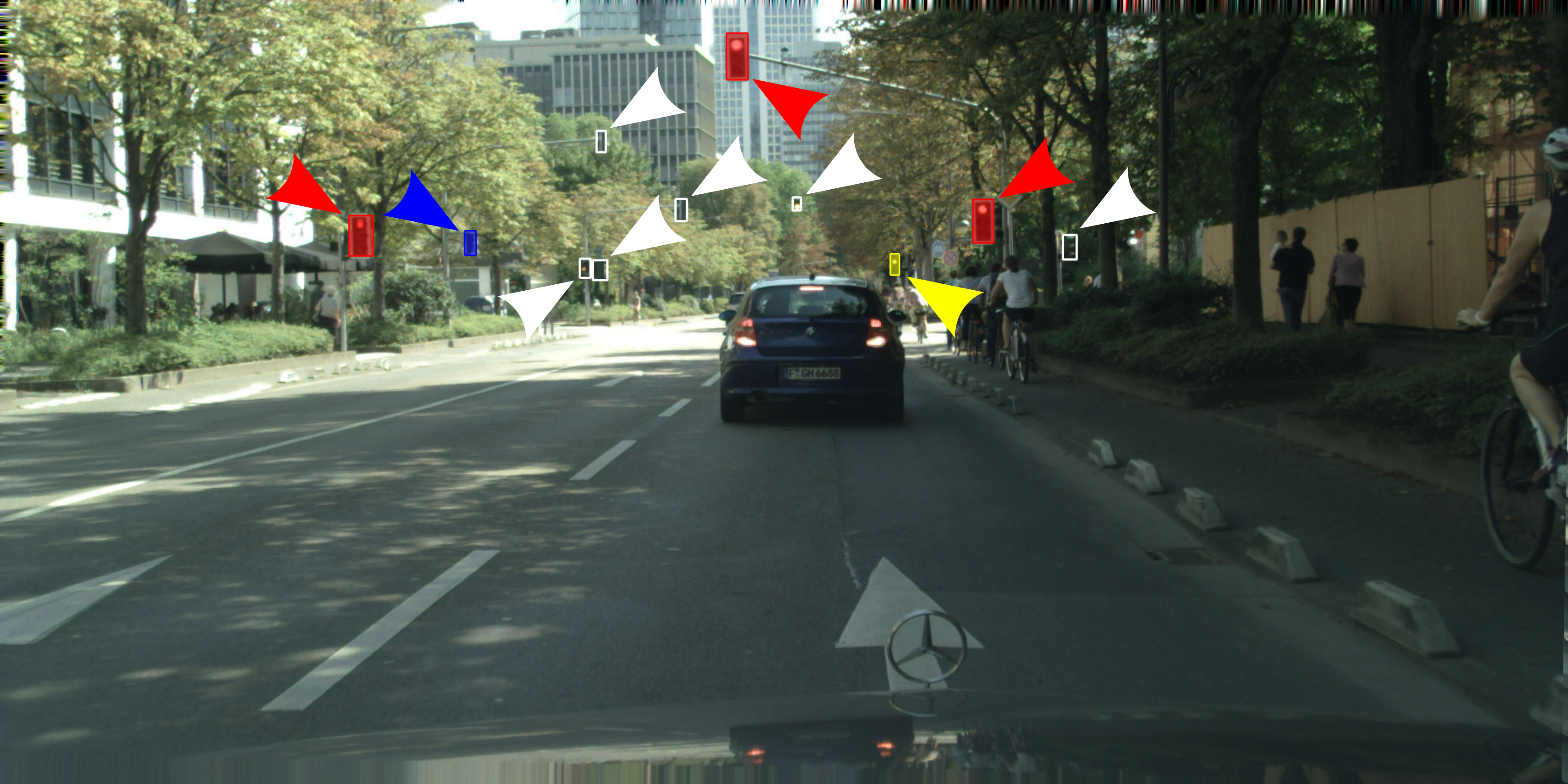}}&
\subfloat{\includegraphics[width = .31\linewidth]{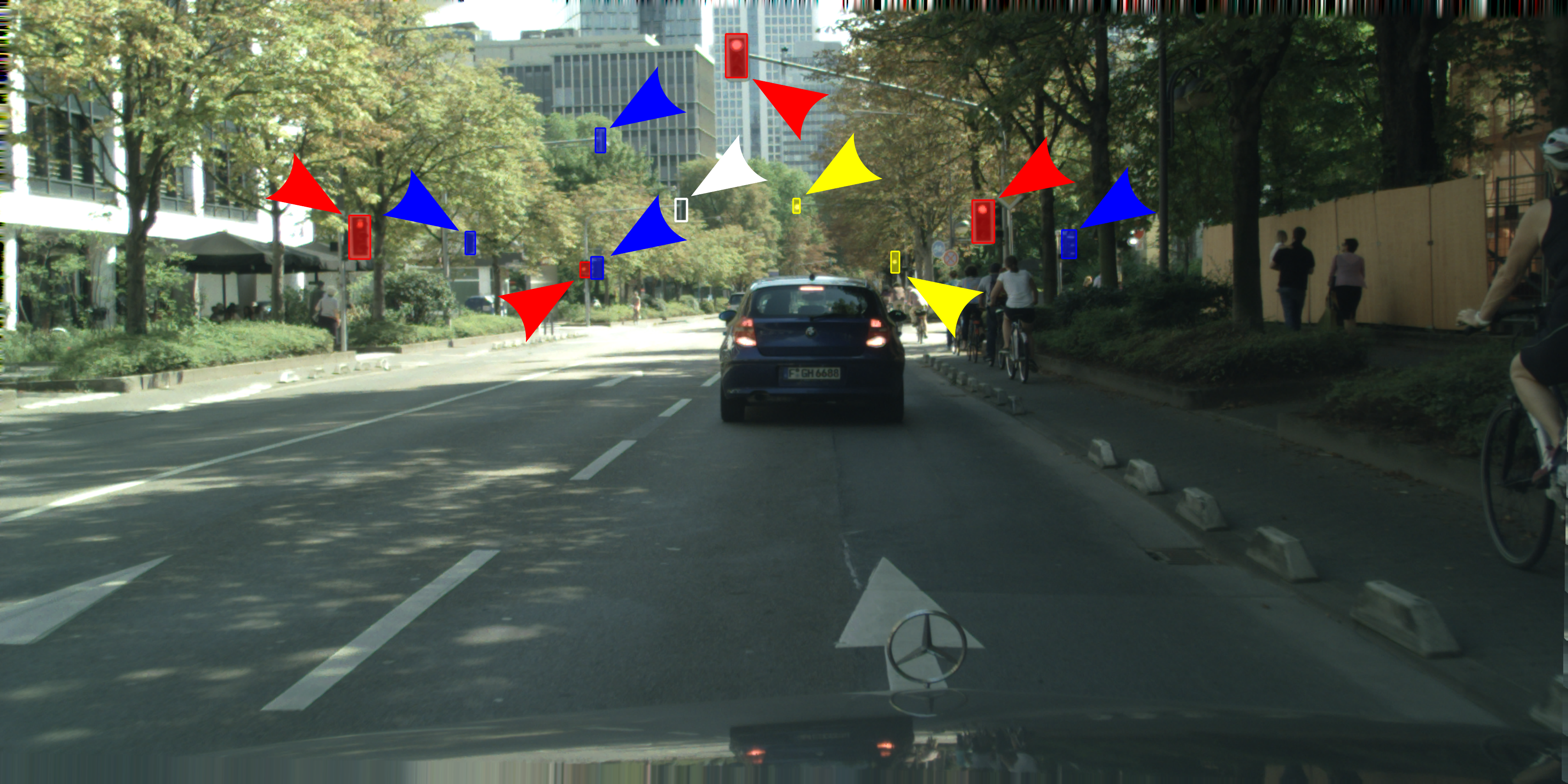}}\\
\subfloat{\includegraphics[width = .31\linewidth]{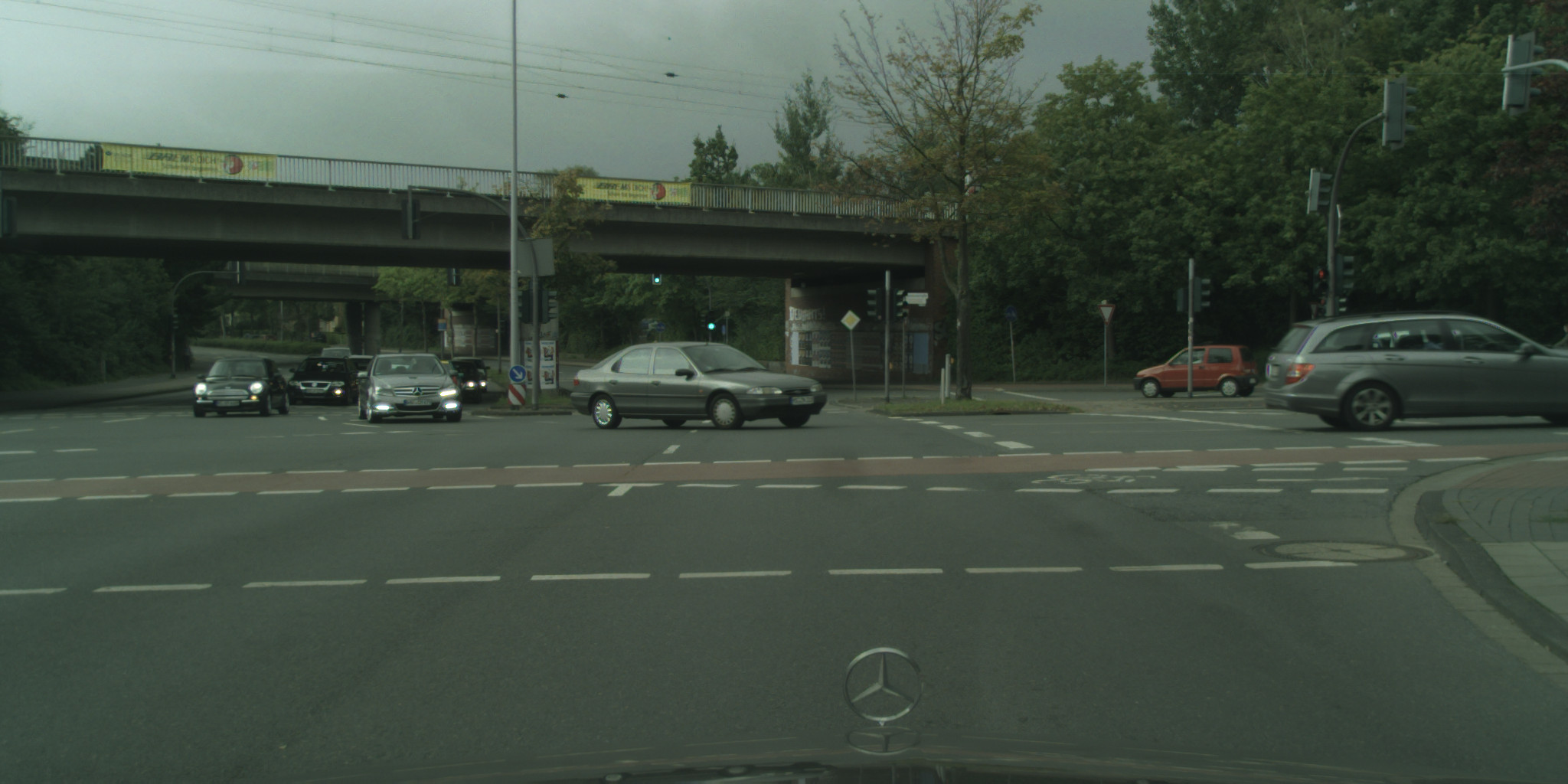}} &
\subfloat{\includegraphics[width = .31\linewidth]{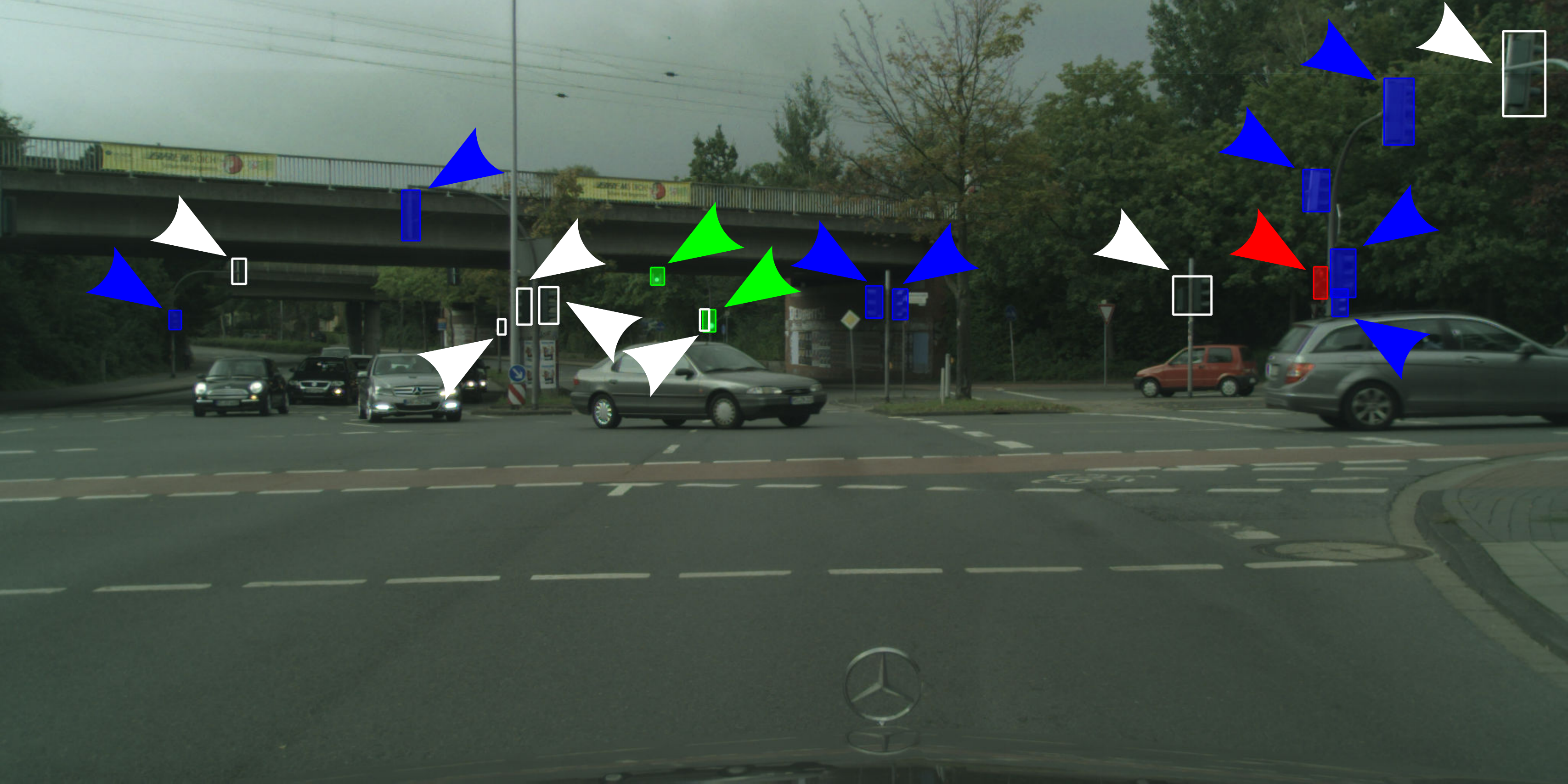}} &
\subfloat{\includegraphics[width = .31\linewidth]{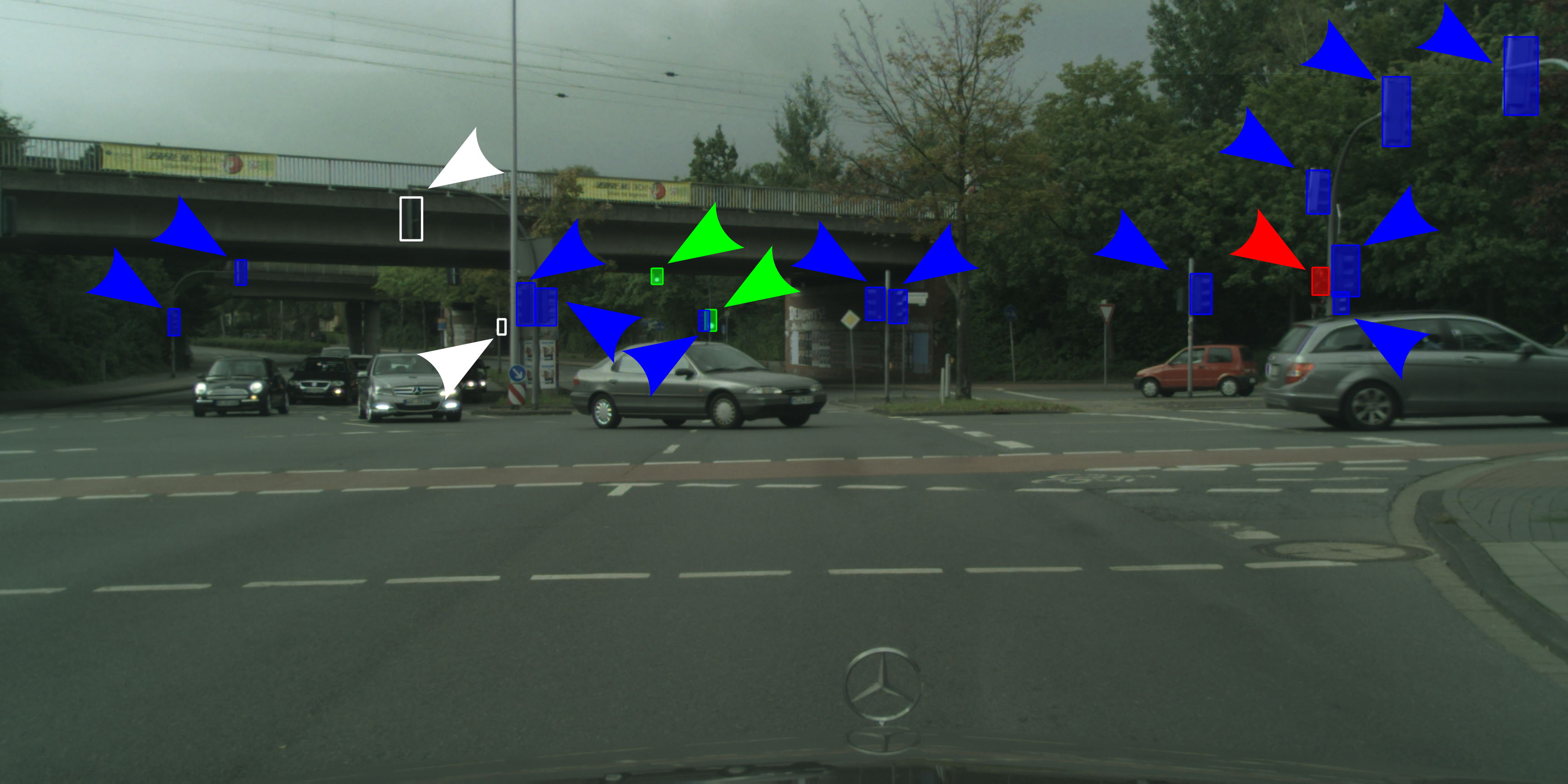}}\\
Image & Faster R-CNN+FPN & Ours
\end{tabular}
\caption{Qualitative results of the Faster R-CNN+FPN~\cite{ren2015faster,lin2017feature} baseline and our proposed traffic light detection system on the CS-TL test set. The color indicates the assigned traffic light state. Blue denotes the state \textit{unknown}, while white denotes missed or misclassified traffic lights. Arrows highlight traffic lights for better visibility.}
\label{fig:qualResCS}
\end{figure}

\subsection{Qualitative Results}
\label{sec:evalQual}
The qualitative results in Fig.~\ref{fig:qualResDriveU} on DTLD and in Fig.~\ref{fig:qualResCS} on the CS-TL dataset support the quantitative results. Comparing the results of TL-SSD and our system on DTLD in Fig.~\ref{fig:qualResDriveU}, it is clearly visible that both systems detect most larger traffic lights. However, the small and tiny traffic lights are only regularly detected by our system. TL-SSD is only able to detect one smaller traffic light, while our system detects almost all of them in the presented examples.

Switching to the recently published CS-TL dataset, Fig.~\ref{fig:qualResCS} shows a comparison of our results and the strong Faster R-CNN+FPN baseline. The results across the four images again show a strong performance of both systems on larger traffic lights, with few exceptions like in the complex example in the final row. Investigating the results in the first two rows in more detail, it is visible that despite a rather low scene complexity, only our system is able to consistently detect smaller traffic lights like the two traffic lights with unknown state (blue) in the second row. The lower two rows show more complex scenes with several traffic lights. Despite the complexity, our system again detects most traffic lights and outperforms Faster R-CNN+FPN, especially on smaller instances. Still, some traffic lights are missed by our system due to low contrast~(see last row).

Overall, these results correspond well to the findings in Tab.~\ref{tab:quanResultsSize}. Our system shows a stronger performance on small and tiny traffic lights compared to other systems, while generally accomplishing very good overall results.

\subsection{Ablation Studies}
\label{sec:evalAbl}
This section investigates the choice of the traffic light proposal generator and the detection head. Both ablation studies were conducted on BSTLD.

\subsubsection{Proposal Generator}
 In Sec.~\ref{sec:method_TLPG}, we proposed a new traffic light proposal generator for locating possible traffic lights. To show the benefit of our proposal generator, which is explicitly designed to discover small and tiny traffic lights, we compare it to two variations of AttentionMask~\cite{WilmsFrintropACCV2018}. AttentionMask is a general-purpose object proposal generator designed to discover small objects. 
 The differences between the variations and our system are the spatial resolution of the most fine-grained feature map in the feature pyramid and the usage of an FPN-based backbone~(see Sec.~\ref{sec:method_TLPG}). While the two versions of AttentionMask utilize a feature map with a downscale factor of 8~(AttentionMask$_8^{128}$~\cite{WilmsFrintropACCV2018}) and 4~(AttentionMask$_4^{96}$~\cite{wilms2022localizing}) as the base of the feature pyramid, we use a feature map with downscale factor 2 as a result of the FPN-based backbone. Note that AttentionMask$_8^{128}$ and AttentionMask$_4^{96}$ do not use an FPN-based backbone. The results in Tab.~\ref{tab:ablProposals} on BSTLD in terms of Average Recall\footnote{Average Recall assesses how many annotated traffic lights are recalled and how precisely they are located, given a specified number of proposals.} (AR) for 1000 and 5000 proposals show that we strongly outperform original AttentionMask with an improvement of $150\%$ on AR@5000. Compared to the variation AttentionMask$_4^{96}$, the improvement is still $12.5\%$. Therefore, our traffic light proposal generator based on the extended FPN-based feature pyramid outperforms general-purpose object proposal generators on the traffic light localization task.

\begin{table}[tb]
\parbox{.56\linewidth}{
\centering
\caption{Traffic light proposal generation results for three proposal generators on BSTLD in terms of AR for 1000 and 5000 proposals.}
\label{tab:ablProposals}
\begin{tabular}{@{}lcc@{}}
\toprule
Proposal Generator & AR@1000 & AR@5000 \\ \midrule
AttentionMask$_8^{128}$~\cite{WilmsFrintropACCV2018} & 0.180 & 0.180 \\ 
AttentionMask$_4^{96}$~\cite{WilmsFrintropACCV2018,wilms2022localizing} & \textbf{0.390} & 0.400 \\
Ours & \textbf{0.390} & \textbf{0.450} \\ \bottomrule
\end{tabular}
}
\hfill
\parbox{.40\linewidth}{
\centering
\caption{Traffic light detection results for three detection heads on BSTLD in terms of mAP.}
\label{tab:ablHead}
\begin{tabular}{@{}lc@{}}
\toprule
Head Structure & mAP \\ \midrule
$2 \times 1024$~\cite{ren2015faster} & 0.554 \\ 
$4 \times 2048$ (ours)\ \ \  & \textbf{0.563} \\ 
$5 \times 2048$ & 0.546 \\ 
 \bottomrule
\end{tabular}
}
\end{table}


\subsubsection{Detection Head}
The detection head in our traffic light detection module refines the traffic light proposals and assigns them a traffic light state. As discussed in Sec.~\ref{sec:method_TLDM}, our new detection head comprises four fully-connected layers with 2048 neurons each. We compare this design to the detection head of Faster R-CNN~\cite{ren2015faster}, which has two layers with 1024 neurons, and a larger detection head with five layers and 2048 neurons each. The results in terms of mAP on BSTLD in Tab.~\ref{tab:ablHead} indicate that our new detection head outperforms Faster R-CNN's detection head ($+1.6\%$) as well as the larger detection head ($+3.1\%$). Hence, our detection head is a good choice for the traffic light detection task.


\section{Conclusion}
In this paper, we addressed the problem of traffic light detection. We specifically focused on the challenging detection of small and tiny traffic lights, which are important for safe, efficient, and eco-friendly driving. Our approach consists of (i) a novel traffic light proposal generator combining the one-shot approach from object proposal generation with fine-grained multi-scale features as well as attention, and (ii) a detection module featuring a new traffic light detection head. The extensive evaluation across three datasets and six methods clearly shows the strong performance of our novel system on small and tiny traffic lights~(at least $+12.6\%$), as well as a strong overall performance on traffic lights of all sizes. Thus, our system can improve safe, efficient, and eco-friendly driving.

%
%
%
\bibliographystyle{splncs04}
\bibliography{lit}

\end{document}